\documentclass{article} 
\usepackage{iclr2027_conference,times}
\usepackage[OT1]{fontenc} 

\usepackage{amsmath,amsfonts,bm}

\def\eqref#1{equation~\ref{#1}}

\def\1{\bm{1}}

\DeclareMathAlphabet{\mathsfit}{\encodingdefault}{\sfdefault}{m}{sl}
\SetMathAlphabet{\mathsfit}{bold}{\encodingdefault}{\sfdefault}{bx}{n}

\usepackage{amsmath}
\usepackage{amssymb}
\usepackage{booktabs}
\usepackage{graphicx}
\usepackage{placeins}
\usepackage{microtype}
\usepackage{xcolor}
\usepackage{hyperref}
\usepackage{url}
\usepackage{wrapfig}
\usepackage{needspace}
\usepackage{tabularx}
\usepackage{textcomp}

\title{SoftSkill: Behavioral Compression\\for Contextual Adaptation}

\newcommand{\hkuaff}{\ensuremath{\spadesuit}}
\newcommand{\huaweiaff}{\ensuremath{\heartsuit}}
\newcommand{\cityuaff}{\ensuremath{\clubsuit}}
\newcommand{\hustaff}{\ensuremath{\diamondsuit}}

\author{
Xijia Tao\textsuperscript{\hkuaff,*} \quad
Yihua Teng\textsuperscript{\huaweiaff,*} \quad
Xinyu Fu\textsuperscript{\huaweiaff,*} \quad
Ziru Liu\textsuperscript{\huaweiaff} \quad
Kecheng Chen\textsuperscript{\cityuaff} \\
\bfseries
Yuzhi Zhao\textsuperscript{\hustaff} \quad
Suiyun Zhang\textsuperscript{\huaweiaff} \quad
Rui Liu\textsuperscript{\huaweiaff,\textdagger} \quad
Lingpeng Kong\textsuperscript{\hkuaff,\textdagger} \\
\textsuperscript{\hkuaff}The University of Hong Kong \quad
\textsuperscript{\huaweiaff}Huawei Research \quad
\textsuperscript{\cityuaff}City University of Hong Kong \\
\textsuperscript{\hustaff}Huazhong University of Science and Technology \\
\textsuperscript{*}Equal contribution \quad
\textsuperscript{\textdagger}Corresponding authors
}

\definecolor{methodpurple}{HTML}{8B32E8}
\definecolor{methodgreen}{HTML}{00A35C}

\newcommand{\method}{\textcolor{methodpurple}{\textsc{SoftSkill}}}
\newcommand{\skill}{s}

\newcommand{\target}{f}
\newcommand{\harness}{\mathcal{H}}
\newcommand{\dataset}{\mathcal{D}}
\newcommand{\reward}{R}
\newcommand{\tbd}{--}
\newcommand{\best}[1]{\textbf{\boldmath #1}}
\newcommand{\acc}[2]{#1{\scriptsize\, (#2)}}
\newcommand{\actionopen}{\texttt{\textless action\textgreater}}
\newcommand{\actionclose}{\texttt{\textless/action\textgreater}}
\newcommand{\nothinghappens}{\texttt{Nothing happens.}}

\newif\ifshowcomments
\showcommentsfalse
\newcommand{\xj}[1]{\ifshowcomments\textcolor{magenta}{\bfseries\small [#1 --xj]}\fi}
\newcommand{\gpt}[1]{}

\iclrfinalcopy
\hypersetup{
  pdftitle={SoftSkill: Behavioral Compression for Contextual Adaptation},
  pdfauthor={Xijia Tao, Yihua Teng, Xinyu Fu, Ziru Liu, Kecheng Chen, Yuzhi Zhao, Suiyun Zhang, Rui Liu, Lingpeng Kong}
}

\begin{document}

\maketitle
\fancyhead{}

\begin{abstract}
Natural-language skills let agents reuse task knowledge, but long skill documents
must be reinterpreted on every call.
We introduce \method{}, which converts a skill document into a compact trainable
context by initializing virtual-token embeddings from its text and optimizing an
additive update with next-token prediction while freezing the language model.
On Qwen3.5--4B, a 32-token soft skill improves over no skill by 7.6, 42.1, and
1.3 points on SearchQA, LiveMath, and DocVQA, respectively, and exceeds the
optimized textual skill on SearchQA and LiveMath by 4.5 and 12.5 points while
replacing artifacts containing hundreds to thousands of tokens.
The same interface also improves multi-step execution: on Qwen3.6--35B-A3B,
OfficeQA and ALFWorld improve over no skill, while in a controlled
Qwen3.5--4B ALFWorld setting, training raises success from 44/134 to 91/134.
Mechanistic interventions show that this gain is carried by a small,
task-aligned continuous displacement: moving halfway along the learned direction
recovers most of the improvement, an equal-norm random displacement does not,
vocabulary projection removes the gain, and early-layer state replacement
strongly suppresses it.
These results show that task supervision can compile a readable skill artifact
into a compact continuous control for both answers and actions without updating
the backbone.
Project page: \url{https://xijia-tao.github.io/softskill}
\end{abstract}


\section{Introduction}

Skill files teach agents how to answer questions, consult evidence, and act in environments. Systems increasingly package this knowledge as reusable Markdown documents, verbal memories, and executable resources~\citep{shinn2023reflexion,zhao2023expel,wang2023voyager,zhang2025agentskills,jiang2026sok,yang2026skillopt}. These artifacts are inspectable and editable, but the model must reinterpret them every time they are loaded, often spending substantial context on prose rather than the resulting behavior.

We ask whether skill text can instead initialize a continuous context shaped by task feedback. Unlike a textual skill, this object can move through embedding space toward directions the target model uses; unlike a weight adapter, it leaves the backbone unchanged and enters through the context interface.

\begin{figure}[t]
\begin{center}
\includegraphics[width=0.98\linewidth]{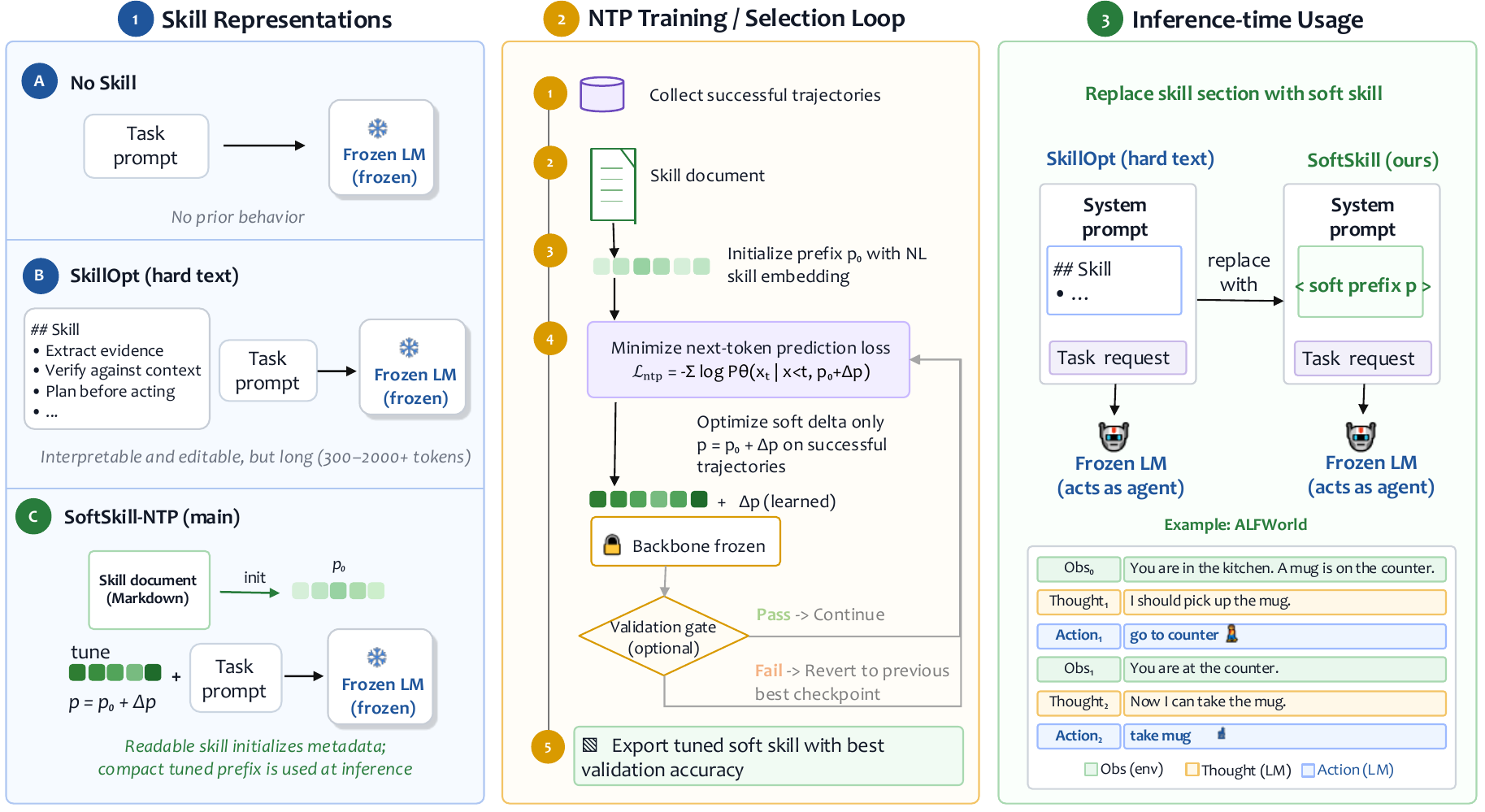}
\end{center}
\caption{\method{} initializes virtual tokens from skill text, tunes a soft-skill update with next-token prediction, and selects the deployed checkpoint using held-out task performance.}
\label{fig:teaser}
\end{figure}

We introduce \method{}, which converts a natural-language skill into initial virtual-token embeddings $p_0$ and learns an additive update $\Delta p$. Next-token prediction uses ground-truth answers for question answering and successful teacher trajectories for agent execution; held-out task performance selects the checkpoint. The soft skill can occupy the prompt's skill section or another supported insertion point. Figure~\ref{fig:teaser} summarizes the workflow.

Our contributions are threefold. \textbf{First}, \method{} provides a frozen-backbone interface that turns readable skill text into a trainable continuous skill block. \textbf{Second}, 32-token skills outperform optimized Markdown skills by 4.5 points on SearchQA and 12.5 on LiveMath, improve DocVQA over no skill, and replace artifacts containing hundreds to thousands of tokens; the same interface improves OfficeQA and ALFWorld. \textbf{Third}, controlled ALFWorld experiments raise success from 44/134 to 91/134 and show that the learned skill changes action dynamics. Continuous-path and layer interventions further identify a small learned displacement that is causally read by the frozen model.


\section{Related Work}
\label{sec:related}

Textual skill systems store reusable instructions that can be inspected, edited, and loaded into an agent's context~\citep{zhang2025agentskills,jiang2026sok,ling2026agentskills}. SkillOpt, TextGrad, and GEPA improve such text using rollout feedback or prompt search~\citep{yang2026skillopt,yuksekgonul2024textgrad,agrawal2025gepa}. Their output remains a readable artifact interpreted at inference time. \method{} instead uses that artifact to initialize embeddings and trains the conditioning representation itself.

Continuous conditioning includes prompt tuning, prefix tuning, learned task vectors, and skill neologisms~\citep{lester2021power,li2021prefix,yang2026taskvectors,berthon2026skillneologisms}. Skill neologisms learn vocabulary-level skill tokens and study continual learning and composition, while KV-Skill injects factorized residual-stream operators without consuming prompt positions~\citep{han2026kvskill}. \method{} specifically converts a reusable natural-language artifact into an input-side trainable skill representation and studies its compression of QA skills and execution in agents. Extended comparisons with prompt search, distillation, and model-native controls appear in Appendix~\ref{sec:appendix-related}.

\section{Method}
\label{sec:method}

\subsection{Problem Setup}

\method{} turns a written skill into a trainable input representation for a frozen model. Let $\target$ be the language model and $\harness$ its execution harness. A task instance $x \in \dataset$ produces a trajectory $\tau = \harness(\target, c(x,p))$ and a task score $\reward(\tau)$. Here $c(x,p)$ is the task context with a block $p$ of $m$ virtual token embeddings inserted at a configured position. A Markdown skill document $\skill$ initializes the block; the deployed context replaces that document with $p$. We parameterize that block as
\[
    p = p_0 + \Delta p,
\]
where $p_0$ comes from skill text and only $\Delta p$ is optimized. The backbone and execution harness remain fixed. Training uses next-token prediction (NTP) on successful behavior, while held-out task reward selects the checkpoint used at test time.

For single-round question answering, the target $y$ is the ground-truth answer, without a chain-of-thought target. The objective is
\begin{equation}
    \Delta p^\star
    =
    \arg\min_{\Delta p}
    \; \mathbb{E}_{(x,y) \sim \dataset_{\mathrm{train}}}
    \left[-\sum_{t=1}^{T}\log p_{\target}(y_t \mid y_{<t}, c(x,p_0+\Delta p))\right],
    \label{eq:ntp}
\end{equation}
with frozen backbone weights. For multi-turn tasks, the same objective is applied to successful teacher assistant decisions along a trajectory, conditioning each decision on the interaction history available at that step. The supervised text can include the full assistant response or the action that is actually executed; Section~\ref{sec:experiments} examines this choice in a controlled ALFWorld study.

This parameterization separates the authored artifact from the behavior learned for a particular target model. The initialization $p_0$ retains the token order and local structure of the source skill, while $\Delta p$ can express directions that do not correspond to discrete vocabulary items. At deployment, the block $p_0+\Delta p$ replaces the source skill text in the configured slot; it is not concatenated with that text. The resulting trainable state consists only of $m$ input vectors, and gradients never update the embedding table or any transformer weight.

\paragraph{Validation-selected optimization.}
We evaluate the initialization and each epoch checkpoint using task reward on a held-out validation split. This is important because token-level NTP loss measures imitation of the supervised target, whereas deployment quality is determined by the benchmark harness after generation. The selected checkpoint is then evaluated once on the held-out test split. Thus, NTP supplies a dense optimization signal, but model selection remains tied to the answer or action behavior that the skill is intended to induce.

\subsection{Initialization and Insertion}

The default initialization takes the first $m$ target-model embeddings of tokenized skill text; the auto-length setting uses the full tokenized artifact. Mean-pooled skill embeddings provide a text-averaged initialization baseline. The insertion slot is configurable: \texttt{skill\_section} places the block where the SkillOpt template expects a skill, and \texttt{prompt\_start} places it before the system prompt. The method therefore supports semantic skill slots rather than requiring every block to precede the full prompt.

SearchQA and LiveMath use \texttt{skill\_section} for the headline comparison; the headline DocVQA result uses \texttt{prompt\_start}, with true skill-section runs reported separately (Appendix~\ref{sec:appendix-docvqa-placement}). The original Markdown can remain as metadata for authoring, retrieval, and provenance even though inference injects only the continuous block. We measure behavioral compression by the skill-context budget needed to achieve task performance; the auto-length agent studies instead hold the source artifact's context budget fixed and ask whether optimization improves its behavior.

\section{Experiments}
\label{sec:experiments}

\subsection{Benchmarks and Models}

We evaluate six low-data benchmarks from the SkillOpt suite~\citep{yang2026skillopt}. SearchQA~\citep{dunn2017searchqa} asks questions over retrieved evidence, LiveMath~\citep{he2026livemathematicianbenchlivebenchmarkmathematicianlevel} tests mathematical problem solving, and DocVQA~\citep{mathew2021docvqa} asks questions about document images. OfficeQA~\citep{officeqa} searches office documents, SpreadsheetBench~\citep{ma2024spreadsheetbenchchallengingrealworld} requires spreadsheet operations, and ALFWorld~\citep{shridhar2021alfworld} tests multi-step household actions. We follow the released splits with validation selection and held-out test reporting. Target models are open-weight Qwen models~\citep{yang2025qwen3} with frozen backbones.

The QA tasks train on ground-truth answers without chain-of-thought targets. The headline comparison uses a 32-token Qwen3.5--4B block: \texttt{skill\_section} for SearchQA and LiveMath and \texttt{prompt\_start} for DocVQA. We also evaluate true DocVQA skill-section insertion (Appendix~\ref{sec:appendix-docvqa-placement}).

The execution tasks train on successful GPT-5.5 teacher trajectories and initialize from a SkillOpt artifact; GPT-5.5 supplies supervision rather than serving as the target model. OfficeQA and SpreadsheetBench use NTP over full assistant responses. Across all six tasks, the nominal train total is 711 instances, including 35 for LiveMath; Appendix Tables~\ref{tab:data-splits} and~\ref{tab:task-status} give splits, usable trajectory counts, and supervision details.

For the controlled Qwen3.5--4B ALFWorld study, we train on the same 294 teacher decisions and evaluate with forced-action decoding. We compare full-response NTP, action-only NTP, and a masked-rationale diagnostic. This 2806-token forced-action protocol is distinct from both the 35B free-form agent comparison and the 32-token QA setting; complete optimization, target-construction, and decoding details are in Appendix~\ref{sec:appendix-alfworld-4b}.

\subsection{Metrics and Baselines}

We select checkpoints by validation task score and report held-out accuracy or success under each benchmark harness. Principal comparisons are no skill, hard skills, SkillOpt, and LoRA; Table~\ref{tab:main-results} also includes LLM skill, Trace2Skill, and GEPA. Human skill, TextGrad, and the full taxonomy appear in Appendix Tables~\ref{tab:baselines} and~\ref{tab:main-results-full}. Reproduction details and differences from the original SkillOpt serving setup are documented in Appendix~\ref{sec:appendix-additional-results}.

\section{Main Results}

\paragraph{Single-round tasks.}
A 32-token \method{} improves Qwen3.5--4B on all three QA tasks while tuning only the skill embeddings (Table~\ref{tab:main-results}). It exceeds SkillOpt on SearchQA ($75.7\pm1.9$ vs.\ $71.2$) and LiveMath ($64.5\pm2.4$ vs.\ $52.0$), and improves over no skill by $7.6$ and $42.1$ points. It also exceeds LLM skill, Trace2Skill, and GEPA on both tasks. On DocVQA, the headline prompt-start result reaches $88.2\pm1.4$ versus $86.9$ without a skill; true skill-section insertion reaches $89.9\pm0.4$ (Table~\ref{tab:placement}). LoRA leads SearchQA at $78.6$. Human skill and TextGrad appear in Appendix Table~\ref{tab:main-results-full}.

\begin{table}[ht]
\caption{Main single-round QA comparison on Qwen3.5--4B. \method{} reports mean $\pm$ standard deviation over three seeds under the default placement. Deltas are versus no skill. Human skill and TextGrad are in Appendix Table~\ref{tab:main-results-full}.}
\label{tab:main-results}
\begin{center}
\small
\begin{tabular}{lccc}
\toprule
Method & SearchQA & LiveMath & DocVQA \\
\midrule
No skill & \acc{68.1}{+0.0} & \acc{22.4}{+0.0} & \acc{86.9}{+0.0} \\
LLM skill & \acc{65.0}{-3.1} & \acc{28.8}{+6.4} & \acc{88.0}{+1.1} \\
Trace2Skill~\citep{ni2026trace2skilldistilltrajectorylocallessons} & \acc{68.5}{+0.4} & \acc{27.2}{+4.8} & \acc{88.0}{+1.1} \\
GEPA~\citep{agrawal2025gepa} & \acc{68.6}{+0.5} & \acc{28.8}{+6.4} & \acc{85.1}{-1.8} \\
SkillOpt~\citep{yang2026skillopt} & \acc{71.2}{+3.1} & \acc{52.0}{+29.6} & \acc{89.0}{+2.1} \\
LoRA~\citep{hu2021lora} & \best{\acc{78.6}{+10.5}} & \acc{58.9}{+36.5} & \acc{87.7}{+0.8} \\
\method{} & \acc{$75.7 \pm 1.9$}{+7.6} & \best{\acc{$64.5 \pm 2.4$}{+42.1}} & \acc{$88.2 \pm 1.4$}{+1.3} \\
\bottomrule
\end{tabular}
\end{center}
\end{table}

\paragraph{Relation to weight adaptation.}
\method{} occupies a different deployment point from LoRA: the backbone remains untouched, and adaptation is carried entirely through the model's input context. The comparison is also empirically complementary. LoRA leads SearchQA at $78.6$ versus \method{}'s $75.7\pm1.9$, while \method{} leads LiveMath at $64.5\pm2.4$ versus $58.9$. On DocVQA, true skill-section insertion reaches $89.9\pm0.4$, above LoRA's $87.7$. These results motivate continuous skill blocks as a compact contextual adaptation interface rather than a claim that one parameter-efficient method dominates every task.

\begin{figure}[t]
\begin{center}
\includegraphics[width=0.95\linewidth]{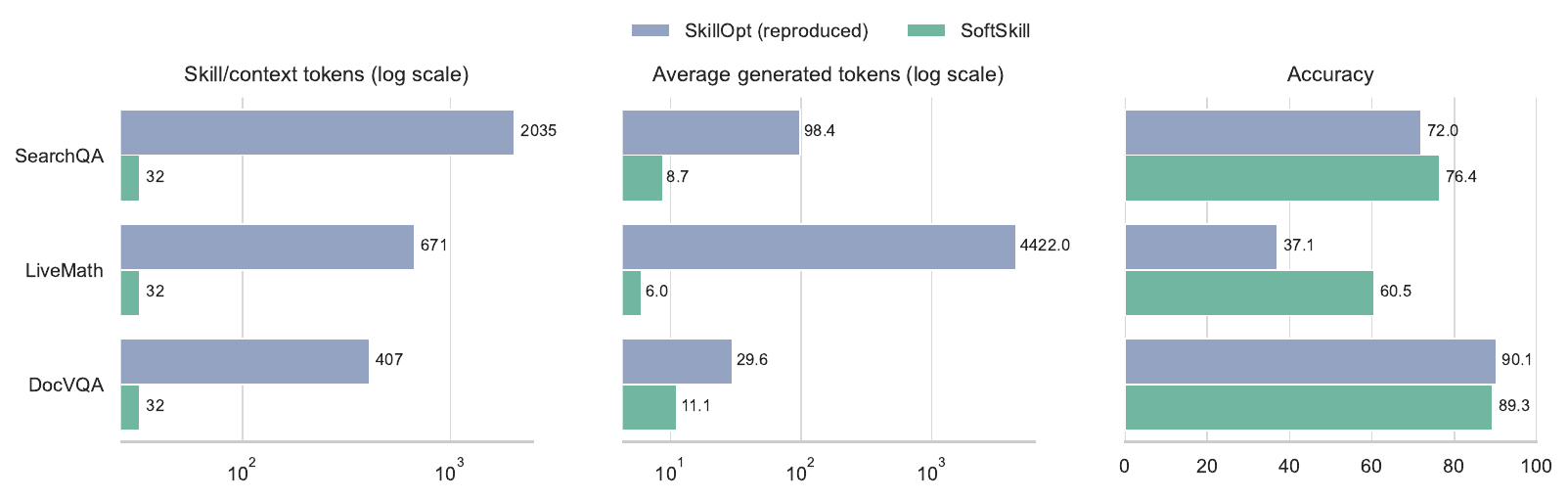}
\end{center}
\caption{Compression diagnostics on Qwen3.5--4B. A 32-token \method{} replaces much longer reproduced SkillOpt artifacts. Output lengths are reported separately because direct-answer NTP can shorten generations without demonstrating compression of reasoning. Token axes use log scale.}
\label{fig:compression-diagnostics}
\end{figure}

\paragraph{Compression diagnostics.}
A 32-token \method{} replaces reproduced SkillOpt artifacts containing 2035, 671, and 407 tokens on SearchQA, LiveMath, and DocVQA while improving the first two tasks (Figure~\ref{fig:compression-diagnostics}). Output lengths fall as well, but that reduction follows the direct-answer targets and does not by itself show compressed reasoning. Reproduction details are in Appendix Table~\ref{tab:appendix-compression-diagnostics}.

\paragraph{Agentic execution.}
In the 35B free-form setting, \method{} improves OfficeQA from 25.6 to 33.8
and ALFWorld from 57.5 to 71.6 over no skill (Table~\ref{tab:agentic-results-placeholder}).
Improvements over the hard skill are task-dependent: OfficeQA improves, whereas
SpreadsheetBench and ALFWorld do not under this training and decoding recipe.
This motivates the controlled ALFWorld study below, which aligns training and
forced-action evaluation to isolate how trajectory supervision changes executable behavior.

\begin{figure}[t]
\centering
\includegraphics[width=\linewidth]{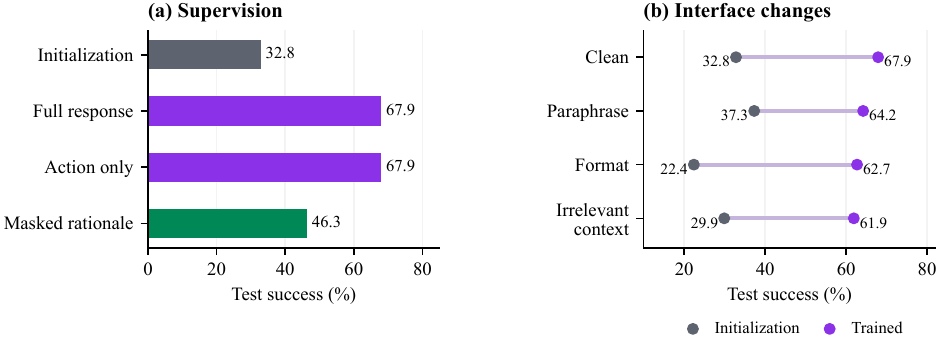}
\caption{Learning and deploying an ALFWorld soft skill on Qwen3.5--4B with forced-action decoding ($n=134$ test games; 2806 skill tokens). (a) Full-response and action-only supervision both reach $91/134$ successes. (b) The action-only checkpoint retains its advantage over initialization under three interface changes, without retraining. Exact values, validation results, and gains appear in Appendix Tables~\ref{tab:alfworld-4b-forced} and~\ref{tab:alfworld-robustness}.}
\label{fig:alfworld-learning}
\end{figure}

ALFWorld provides a controlled test of how trajectory supervision becomes an executable skill. On Qwen3.5--4B, we train on the same 294 teacher decisions and evaluate using forced-action decoding. Full-response and action-only NTP each raise test success from $44/134$ to $91/134$ (Figure~\ref{fig:alfworld-learning}a). A masked-rationale diagnostic, which conditions actions on a gold rationale unavailable at inference, reaches $62/134$. Optimization and decoding details are in Appendix~\ref{sec:appendix-alfworld-4b}.

The equal aggregate result does not mean that full-response and action-only skills induce identical behavior: they disagree on 26 of the 134 games. Instead, it shows that rationale-token loss is unnecessary for the measured gain when the executed action is predicted under the same context used at deployment. The masked-rationale condition isolates the converse mismatch. Its action-token loss is small, but the action is learned after a teacher-forced rationale that forced-action evaluation never provides. Matching the conditioning context, rather than merely choosing which target tokens receive loss, determines whether the supervised action transfers to rollout.

\begin{table}[t]
\caption{
Agentic-task comparison under the 35B free-form setting.
\method{} is selected by validation; improvements over the hard skill are task-dependent in this setting.
The complete length grid is in Appendix Table~\ref{tab:agentic-ablation-grid}.
}
\label{tab:agentic-results-placeholder}
\begin{center}
\small
\begin{tabular}{lccc}
\toprule
Task & No skill & Hard final & Val-selected \method{} \\
\midrule
OfficeQA & 25.6 & 26.7 & \best{33.8} (len=32) \\
Spreadsheet & 39.6 & \best{52.5} & 28.2 (len=8) \\
ALFWorld & 57.5 & \best{79.1} & 71.6 (auto) \\
\bottomrule
\end{tabular}
\end{center}
\end{table}

\paragraph{Robustness to interface perturbations.}
The learned skill retains its advantage without retraining under instruction paraphrasing, formatting changes, and irrelevant context (Figure~\ref{fig:alfworld-learning}b). Success remains $61.9$--$64.2\%$ under these perturbations versus $67.9\%$ on the original interface, with gains of $26.9$--$40.3$ points over initialization ($p<10^{-8}$ for every paired comparison; Appendix Table~\ref{tab:alfworld-robustness-paired}).

These comparisons keep the games, legal actions, and decoding protocol fixed while changing how instructions and context are presented. Across the three interfaces, training converts 38--58 initialization failures into successes and reverses only 1--4 initialization successes. The persistent paired advantage indicates that the learned action bias is not tied only to the exact wording or formatting of the training prompt.

\paragraph{What changes in the learned agent behavior?}
The success gain is accompanied by shorter trajectories, fewer loops, and fewer rejected actions. Table~\ref{tab:trajectory-compact} gives the strongest paired statistics; Appendix~\ref{sec:appendix-alfworld-behavior} provides complete definitions and subgroup results. Section~\ref{sec:softskill-mechanism} then examines the learned update itself.

\section{Analysis}

We next ask how many embeddings the QA skills need, how much optimization contributes beyond text initialization, and how the frozen model reads the learned update.

\paragraph{A small skill is sufficient for the measured QA behavior.}
Adding embeddings beyond length 32 yields no consistent gain (Appendix Table~\ref{tab:prefix-length-hard}). Placement matters more on LiveMath: at length 32, \texttt{skill\_section} reaches $64.5\pm2.4$ versus $50.0\pm9.1$ at \texttt{prompt\_start} (Table~\ref{tab:placement}). SearchQA changes little across placements. On DocVQA, true \texttt{skill\_section} insertion reaches $89.9\pm0.4$, compared with the headline prompt-start result of $88.2\pm1.4$. These results support 32 tokens as a practical default while retaining both insertion positions as method choices.

\begin{table}[ht]
\caption{Insertion placement for the default NL-initialized length-32 skill block. Mean $\pm$ standard deviation over three seeds. Scores are test-hard accuracy; deltas are versus no skill.}
\label{tab:placement}
\begin{center}
\begin{tabular}{lccc}
\toprule
Placement & SearchQA & LiveMath & DocVQA \\
\midrule
\texttt{prompt\_start} & \acc{$76.4 \pm 0.0$}{+8.3} & \acc{$50.0 \pm 9.1$}{+27.6} & \acc{$88.2 \pm 1.4$}{+1.3} \\
\texttt{skill\_section} & \acc{$75.7 \pm 1.9$}{+7.6} & \acc{$64.5 \pm 2.4$}{+42.1} & \acc{$89.9 \pm 0.4$}{+3.0} \\
\bottomrule
\end{tabular}
\end{center}
\end{table}

\paragraph{Insertion changes how the model encounters the skill.}
The two placements expose the same learned object through different prompt structure. \texttt{prompt\_start} presents the block before the system prompt, whereas \texttt{skill\_section} places it where the task template normally carries procedural guidance. SearchQA is nearly invariant to this choice, but LiveMath gains $14.5$ points from the skill-section placement in the three-seed mean. DocVQA also improves from $88.2$ to $89.9$ when the block is inserted into the true skill slot. This task dependence explains why \method{} treats insertion as part of the deployment interface rather than defining a soft skill as a prefix that must always occur first.

\paragraph{Compression is behavioral rather than lexical.}
The 32-token block is not required to reconstruct the source Markdown or preserve its token-level semantics. Its role is to induce task behavior under the frozen model and harness. This distinction is visible in three results: the short block improves SearchQA and LiveMath while replacing much longer artifacts; mean-pooled embeddings can be ineffective before optimization yet strong afterward; and the ALFWorld skill loses its trained behavior when snapped back to vocabulary embeddings even though every nearest token is unchanged by training. Together, these results identify the compressed object as a learned contextual control initialized by text, rather than a discrete summary of that text. The original document remains useful as an authorable artifact and provenance record.

\paragraph{Optimization turns an initialization into a skill.}
Optimization contributes substantial behavior beyond inserting text-derived vectors (Table~\ref{tab:init-delta}). Mean-pooled embeddings score $0.8$ on LiveMath and $0.0$ on DocVQA before training, yet reach $59.7$ and $89.6$ after the same NTP recipe. Thus, a starting point that does not itself produce useful generations can still support a strong learned skill.

Initialization nevertheless shapes the optimization path. Natural-language initialization is strongest on SearchQA, reaching $77.3$ with skill-section insertion, while SkillOpt-artifact initialization reaches the best LiveMath result at $66.1$. On DocVQA, several trained initializations cluster near $89$ despite sharply different untrained behavior. No single source dominates every task or placement. The text supplies a structured $p_0$; validation-selected $\Delta p$ turns it into the deployed control. Full seed and placement provenance is reported in Appendix~\ref{sec:appendix-docvqa-placement}.

\begin{table}[ht]
\caption{Initialization sensitivity. Trained rows use the same NTP recipe with different $p_0$ values and validation-selected checkpoints; the untrained row measures mean-pooled embeddings before optimization.}
\label{tab:init-delta}
\begin{center}
\small
\resizebox{0.95\linewidth}{!}{
\begin{tabular}{llccc}
\toprule
Initialization & Placement & SearchQA & LiveMath & DocVQA \\
\midrule
Reference & No skill & \acc{68.1}{+0.0} & \acc{22.4}{+0.0} & \acc{86.9}{+0.0} \\
Reference & SkillOpt & \acc{71.2}{+3.1} & \acc{52.0}{+29.6} & \acc{89.0}{+2.1} \\
\midrule
NL skill & \texttt{prompt\_start} & \acc{$76.4$}{+8.3} & \acc{$50.0$}{+27.6} & \acc{$88.2$}{+1.3} \\
NL skill & \texttt{skill\_section} & \best{\acc{$77.3$}{+9.2}} & \acc{$64.5$}{+42.1} & \best{\acc{$89.9$}{+3.0}} \\
Mean-pooled (untrained) & \texttt{prompt\_start} & \acc{70.5}{+2.4} & \acc{0.8}{-21.6} & \acc{0.0}{-86.9} \\
Mean-pooled & \texttt{prompt\_start} & \acc{72.6}{+4.5} & \acc{59.7}{+37.3} & \acc{89.6}{+2.7} \\
Mean-pooled & \texttt{skill\_section} & \acc{76.5}{+8.4} & \acc{42.7}{+20.3} & \acc{89.6}{+2.7} \\
SkillOpt artifact & \texttt{prompt\_start} & \acc{76.4}{+8.3} & \best{\acc{66.1}{+43.7}} & \acc{88.5}{+1.6} \\
SkillOpt artifact & \texttt{skill\_section} & \acc{75.8}{+7.7} & \acc{63.7}{+41.3} & \acc{89.0}{+2.1} \\
\bottomrule
\end{tabular}
}
\end{center}
\end{table}

\paragraph{The learned skill changes sequential behavior.}
The success gain in Figure~\ref{fig:alfworld-learning} corresponds to a broad shift in how the agent acts, not only whether the final goal is reached. Table~\ref{tab:trajectory-compact} summarizes deterministic measurements from the paired forced-action trajectories. Training reduces both explicit environment rejections and repetitive loops, while shortening the median episode from the maximum horizon to eight steps. On the 48 tasks converted from failure to success, the loop rate falls from $64.6\%$ to $4.2\%$. These changes connect successful-action supervision to the model's downstream interaction dynamics; Appendix~\ref{sec:appendix-alfworld-behavior} reports the full metric definitions, subgroup statistics, and trajectory exemplars.

\begin{table}[ht]
\caption{Paired ALFWorld trajectory behavior for initialization and the action-only trained skill ($n=134$).}
\label{tab:trajectory-compact}
\begin{center}
\small
\begin{tabular}{lcc}
\toprule
Metric & Initialization & Trained \\
\midrule
Success & $44/134$ & $\mathbf{91/134}$ \\
Median episode length & 50 & $\mathbf{8}$ \\
Rejected actions / episode & 5.17 & $\mathbf{2.46}$ \\
Episodes containing a loop & $47.8\%$ & $\mathbf{21.6\%}$ \\
Fail$\rightarrow$success subset: loops & $64.6\%$ & $\mathbf{4.2\%}$ \\
\bottomrule
\end{tabular}
\end{center}
\end{table}

\paragraph{Scaling and portability.}
\method{} remains effective as the backbone scales. SearchQA rises from $77.3$ on Qwen3.5--4B to $80.9$ on 9B and $82.3$ on Qwen3.6--35B-A3B; DocVQA rises from $89.3$ to $92.5$ and $93.3$. The latter also retains useful behavior under direct vocabulary-projection transfer: the 4B$\rightarrow$35B block scores $93.0$, close to the target-trained $93.3$, and the reverse transfer scores $90.4$. SearchQA retains a smaller transfer gain, while LiveMath is non-monotonic with scale and requires target-side training in the tested directions. Full scale and transfer results are in Appendix Tables~\ref{tab:scale-transfer}, \ref{tab:model-size-hard}, and~\ref{tab:cross-model-transfer}.

\begin{figure}[t]
\centering
\includegraphics[width=\linewidth]{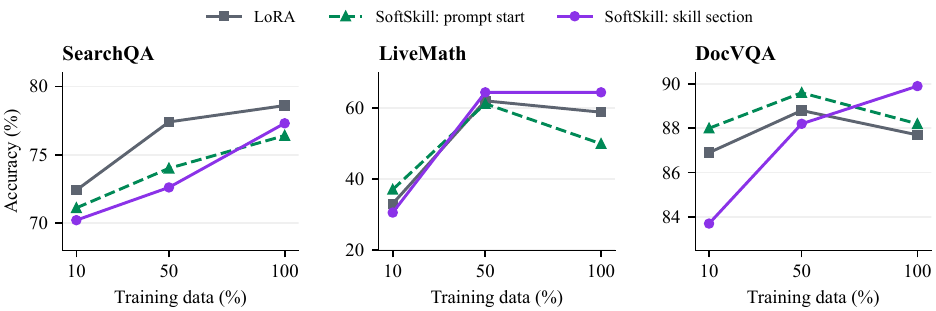}
\caption{Accuracy as training data increases on Qwen3.5--4B. \method{} uses 32 virtual tokens; colors and markers distinguish LoRA and the two insertion locations. Axes use task-specific ranges. Exact values are in Appendix Table~\ref{tab:data-scaling-placeholder}.}
\label{fig:data-scaling}
\end{figure}

\paragraph{Data scaling.}
The same 32-token block remains competitive as the training set shrinks (Figure~\ref{fig:data-scaling}). On LiveMath, \texttt{prompt\_start} reaches $37.1$ with 10\% of the data, above LoRA's $33.1$, and \texttt{skill\_section} reaches $64.5$ at both 50\% and 100\%, above LoRA. SearchQA still favors LoRA at every fraction. Exact values are in Appendix Table~\ref{tab:data-scaling-placeholder}.

\paragraph{Checkpoint selection.}
Lower NTP loss does not reliably select higher task success, motivating validation-based checkpoint selection. Across QA runs, selected checkpoints span all three training epochs; agent runs more often peak after the first epoch. This separation between optimization loss and deployed behavior is especially visible in the masked-rationale ALFWorld diagnostic, where action NLL continues to decrease after validation success has saturated. Appendix Figures~\ref{fig:loss-vs-validation-accuracy} and~\ref{fig:validation-checkpoint-frequency} report the full loss/accuracy and selected-epoch analyses.

\subsection{How does SoftSkill optimization steer a frozen model?}
\label{sec:softskill-mechanism}

\begin{figure}[t]
\centering
\includegraphics[width=\linewidth]{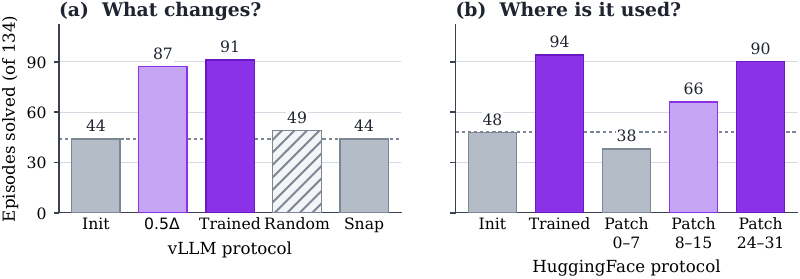}
\caption{\textbf{Mechanistic analysis of an optimized SoftSkill.}
\textbf{Left:} On ALFWorld, half of the learned continuous displacement recovers most of the trained gain, whereas an equal-norm random displacement does not; projecting the trained prefix to its nearest vocabulary embeddings returns exactly to the initialization outcomes.
\textbf{Right:} Under a separate matched HuggingFace inference path, replacing trained prefix states by initialization states in early decoder layers removes the behavioral gain, while the same intervention in late layers has little effect.}
\label{fig:alfworld-mechanism}
\end{figure}

We use the ALFWorld action-only checkpoint as a mechanistic case study of what optimization changes and where the frozen model reads that change (Figure~\ref{fig:alfworld-mechanism}). Geometry and continuous-path tests use the vLLM prompt-embeds evaluation that yields $44/134$ at initialization and $91/134$ after training; the layer intervention uses a matched HuggingFace path described below.

Let $\Delta=p_{\mathrm{trained}}-p_{\mathrm{init}}$. Its Frobenius norm is only $0.109\|p_{\mathrm{init}}\|_F$, and the trained and initialized vectors have mean cosine similarity $0.994$, yet all $2{,}806$ positions move. The displacement is distributed: its leading singular component contains only $3.9\%$ of the energy, and the largest $10\%$ of positions contain $19\%$. Every position nevertheless retains the same nearest vocabulary token after training.

Halfway interpolation, $p_{\mathrm{init}}+0.5\Delta$, already reaches $87/134$ successes, compared with $91/134$ for the trained skill (McNemar $p=0.45$). It converts 43 initialization failures and loses none of the initialization successes. An equal-norm isotropic perturbation reaches only $49/134$ ($p=0.23$ versus initialization), while snapping the trained vectors to their nearest vocabulary embeddings exactly reproduces all $44/134$ initialization outcomes. The gain therefore follows a task-aligned direction in continuous embedding space rather than token substitution or displacement magnitude alone.

The controls separate three explanations that raw accuracy cannot distinguish. Interpolation tests whether behavior emerges smoothly along the learned direction; the equal-norm perturbation tests whether moving away from initialization is sufficient; and vocabulary snapping tests whether optimization merely crosses into different discrete-token neighborhoods. Only the learned direction produces the large gain. Moreover, halfway interpolation recovers most of that gain before reaching the final checkpoint, whereas the snapped endpoint behaves exactly like initialization on every episode. The relevant information therefore resides inside the continuous geometry around the text embeddings.

We next ask where the frozen transformer reads this offset. For each selected layer band, we run matched initialization and trained forwards, replace only the trained prefix positions with their initialization hidden states, and then continue the trained forward through the remaining layers. Task-token states are never directly patched. This intervention uses a separate HuggingFace greedy-decoding path because it requires access to intermediate states; on that path the initialization and trained endpoints are $48/134$ and $94/134$.

Patching layers $0$--$7$ removes the gain ($38/134$), patching layers $8$--$15$ retains part of it ($66/134$), and patching layers $24$--$31$ leaves most of it intact ($90/134$). Relative to the trained endpoint, these interventions change success by $-41.8$, $-20.9$, and $-3.0$ points, respectively. The early patch even falls ten games below initialization, showing that replacing the learned early representation is more disruptive than supplying the initialized prefix end to end.

\begin{wrapfigure}{r}{0.42\textwidth}
\vspace{-0.8em}
\centering
\includegraphics[width=\linewidth]{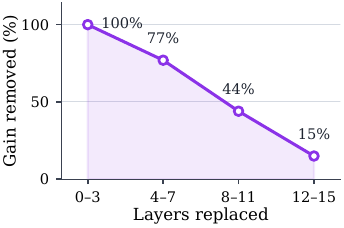}
\caption{Share of the trained-over-initialization action-log-probability gap removed when initialization states replace the trained prefix. This is a teacher-forced probe on training decisions, not an environment rollout.}
\label{fig:alfworld-probe}
\vspace{-0.6em}
\end{wrapfigure}
A finer teacher-forced sweep gives the same trend (\mbox{Figure~\ref{fig:alfworld-probe}}). Replacing layers $0$--$3$, $4$--$7$, $8$--$11$, and $12$--$15$ removes $100\%$, $77\%$, $44\%$, and $15\%$ of the trained-over-initialization action-log-probability gain. The skill information remains in the input block, but the earliest layers are causally necessary for converting its learned offset into action behavior in this checkpoint. Appendix~\ref{sec:appendix-alfworld-mechanism} reports full paired statistics and inference-path details.

\paragraph{Evidence across levels of behavior.}
The experiments connect the learned representation to outcomes at four levels. First, held-out QA accuracy shows that 32 vectors can outperform the long textual skill from which they are initialized. Second, initialization controls show that useful behavior appears after optimizing $\Delta p$: even a mean-pooled start that initially produces almost no correct LiveMath or DocVQA answers becomes effective under NTP. Third, paired ALFWorld rollouts connect the trained block to sequential behavior, with gains persisting across prompt perturbations and coinciding with fewer loops and rejected actions. Finally, continuous-path and hidden-state interventions isolate the learned offset itself. A matched random displacement does not reproduce the gain, vocabulary projection removes it, and early-layer state replacement causally suppresses it.

These levels rule out increasingly weak explanations. The QA result alone could reflect an effective but opaque prompt; the initialization comparison shows that text insertion is insufficient; the trajectory analysis shows that the agent changes how it acts before terminal success; and the interventions show that this change depends on the continuous displacement carried by the skill block. Together they support the paper's central interpretation: task supervision compiles a readable skill artifact into a compact, model-native conditioning object.

\paragraph{Scope.}
The results support a compact task-conditioning interpretation. On QA, 32 vectors are sufficient even when the source artifact contains hundreds or thousands of tokens. On ALFWorld, the same optimization interface changes action selection, loops, and rejected actions, remains effective across prompt perturbations, and yields a measurable continuous offset with an early-layer causal effect. The learned object is therefore more than a replay of its text-derived initialization.

The useful conditioning remains tied to a task and interface: SpreadsheetBench is less favorable under the tested trajectory recipe, and direct cross-model transfer varies by task. This boundary suggests a practical division of labor. Natural-language skills remain the portable, editable source artifact; target-side optimization compiles them into compact controls; and held-out task behavior determines which compiled skill to deploy. Because the backbone stays frozen, multiple task-specific blocks can share one model and be selected at the context boundary without merging weight updates. Extended limitations on serving access, auditability, composition, and transfer appear in Appendix~\ref{sec:appendix-limitations}.

\paragraph{Deployment implications.}
The input-side interface preserves the surrounding application scaffold. A serving system can keep its system message, tool definitions, and task template fixed while replacing a long skill document with the selected embedding block at the same semantic slot. The 32-token QA results show the resulting context savings directly, while the auto-length agent results show that the same interface can refine behavior even when compression is not the immediate goal. Separating the authored and deployed objects also preserves a clear lifecycle: users write and audit natural language, task examples optimize the continuous block, validation chooses a checkpoint, and inference loads only the block associated with the requested skill. This makes \method{} compatible with frozen shared backbones and avoids maintaining a separately modified model for each task.

\FloatBarrier


\FloatBarrier
\begin{samepage}
\section{Conclusion}

\method{} turns a natural-language skill into a trainable continuous context for a frozen model. A 32-token \method{} outperforms optimized textual skills on SearchQA and LiveMath and improves DocVQA over no skill, while agent experiments show gains on OfficeQA and ALFWorld. Controlled ALFWorld experiments further show that task supervision learns a small continuous offset that changes sequential action behavior and is causally read by early transformer layers. These results establish text-derived continuous skill blocks as a compact interface for task adaptation without updating the backbone.
\par
\end{samepage}

\bibliography{references}
\bibliographystyle{iclr2027_conference}

\clearpage
\appendix
\section{Extended Related Work}
\label{sec:appendix-related}

Several lines of work study how language-model behavior can be adapted without full model retraining. Textual skill and prompt-optimization methods modify the artifact that the model reads; soft-prompt, prefix-tuning, skill-neologism, and task-vector methods modify continuous conditioning or activation-level controls; adapter-based methods modify additional model parameters; residual-stream operators such as KV-Skill inject model-native skill interfaces that do not occupy prompt positions; and distillation or feedback-based methods transfer behavior from demonstrations, rollouts, or preferences. Recent self-improvement frameworks also study how multiple adaptation levers can be coordinated: SIA jointly updates an agent harness and model weights in a closed feedback loop~\citep{hebbar2026siaselfimprovingai}. From this perspective, \method{} occupies an input-side interface between harness edits and weight updates: it changes continuous skill block embeddings while keeping both the executable harness and backbone weights fixed. We review related work according to the representation being optimized and the deployment interface it induces.

\paragraph{Reusable skills and agent experience.}
Skill systems package reusable agent behavior as natural-language files that can be selected, inspected, and loaded at inference time~\citep{zhang2025agentskills,ling2026agentskills}. Surveys of the agent-skill layer emphasize progressive disclosure, executable packages, marketplaces, and composition as first-class system problems~\citep{jiang2026sok}. Complementary work studies how skills are generated, mined, or composed rather than how a single skill is compressed into a model-native control~\citep{zhou2026skillgenbench,wang2026skillx,shen2026skillfoundry,zhao2026skillcomposer}. SkillOpt \citep{yang2026skillopt} formulates skill learning as optimization over a single external natural-language state: rollout evidence and held-out validation guide bounded edits to a Markdown skill artifact. Related agent-learning systems store reusable experience in other explicit forms. Voyager builds a library of executable skills for an embodied agent~\citep{wang2023voyager}, while Reflexion and ExpeL convert trial-and-error experience into verbal memory or lessons that can guide later attempts without updating model weights~\citep{shinn2023reflexion,zhao2023expel}. These systems share the premise that useful behavior can be externalized as text, code, or memory. \method{} asks a complementary question: whether some of that reusable behavior can be compressed into a learned continuous context object consumed directly by the target model. We do not evaluate skill retrieval, marketplace routing, or multi-skill composition.

\paragraph{Text-space prompt and instruction optimization.}
Prompt and instruction optimization methods search over textual artifacts rather than model weights. AutoPrompt uses gradient guidance to discover discrete trigger tokens~\citep{shin2020autoprompt}; APE and OPRO use language models to propose and refine natural-language instructions~\citep{zhou2022ape,yang2023opro}; DSPy treats prompts and pipelines as optimizable program components~\citep{khattab2023dspy}. InstructZero, TextGrad, GEPA, and SkillOpt similarly optimize language artifacts with black-box or feedback-driven search~\citep{chen2023instructzero,yuksekgonul2024textgrad,agrawal2025gepa,yang2026skillopt}. These approaches retain the portability and auditability of text, but the target model must still reinterpret the optimized artifact at inference time. \method{} instead optimizes the embedding object that conditions the target model, requiring white-box access during training while keeping the backbone frozen.

\paragraph{Soft prompts, prefix tuning, skill neologisms, and learned task vectors.}
Prompt tuning, prefix tuning, and P-tuning variants show that a small number of continuous virtual tokens can adapt a frozen language model~\citep{lester2021power,li2021prefix,liu2022ptuningv2}. Subsequent work studies sensitivity to initialization, robustness, prompt length, and reparameterized or residual prompt updates~\citep{wang2023infoprompt,razdaibiedina2023residual}. Dynamic Prompt Corruption (DPC) further treats a learned soft prompt as an inference-time object whose instance-specific harmful influence can be detected and selectively masked~\citep{fan2025improvingcomplexreasoningdynamic}. This is closest to our \texttt{prompt\_start} setting with natural-language initialization, but the intervention is complementary: DPC is a post-hoc repair mechanism for an existing prompt, whereas \method{} trains a compact skill block from a skill artifact and evaluates whether it can replace or refine that artifact in deployment. LoRA and related low-rank adaptation methods provide a stronger parameter-efficient tuning baseline when updating model parameters is allowed~\citep{hu2021lora}.

Skill Neologisms~\citep{berthon2026skillneologisms} is a close conceptual neighbor. Both approaches keep the backbone frozen and optimize continuous skill representations with cross-entropy-style supervision. Skill neologisms learn new vocabulary-level soft tokens on skill-centered datasets and focus on compositional continual learning, including zero-shot composition of independently learned skills. \method{} instead starts from a reusable natural-language skill artifact, learns a task-level input skill block from low-data answers or successful trajectories, and evaluates compression relative to long skill artifacts such as SkillOpt. Our experiments evaluate task-level adaptation; decomposition and zero-shot skill composition remain open.

Learned task vectors provide a bridge between prompt tuning and activation-level task representations~\citep{yang2026taskvectors}. Those vectors are typically injected into residual streams at chosen layers and positions, whereas \method{} optimizes an input-side skill block that occupies prompt positions. These interfaces differ in where they intervene in the model.

\paragraph{Model-native skill operators.}
KV-Skill~\citep{han2026kvskill} encodes task knowledge as continuous operators and reports a direct comparison against \method{} on LiveMath. KV-Skill uses external factorized operators read through a learned residual-stream interface rather than consuming prompt positions. It has a text-registration path, which compiles an authored skill into an operator, and a reward-learning path, which discovers a compact operator from outcomes. Across multiple backbones it reports strong results, including 77.2 on Qwen3.5--4B LiveMath for registration versus 64.5 for \method{} in that comparison. \method{} uses a simpler input-side continuous skill block and trains it directly with NTP on answers or successful trajectories. The papers differ in injection site, training objective, and empirical scope. Our study emphasizes behavioral compression of long text skills, the text-initialized skill block deployment path, the low-data SkillOpt comparison, and the aligned agentic loss-support study.

\paragraph{Latent reasoning.}
Recent latent-reasoning work uses continuous states not only as static conditioning vectors, but also as dynamically generated intermediate computations. Monet trains a vision-language model to reason through latent visual states produced inside the reasoning trajectory~\citep{wang2025monetreasoninglatentvisual}. Soft-prompt and prefix-tuning methods learn task-level latent inputs that are fixed before inference. \method{} studies this static conditioning interface for reusable procedural knowledge: the learned object is a persistent skill representation initialized from text and consumed as an input-side skill block, rather than a per-example latent scratchpad generated by the model.

\paragraph{Behavioral compression from supervision.}
Knowledge distillation compresses behavior from a larger or more expensive source into a smaller deployable object~\citep{hinton2015distilling}. Behavior cloning and imitation learning similarly train a policy from demonstrated or successful trajectories~\citep{ross2011dagger}. The main \method{} objective is closer to this supervised view than to reinforcement learning: next-token prediction on answers or successful turns trains the skill block to bias a frozen decoder toward observed successful behavior. On ALFWorld we ablate whether that loss must cover rationale tokens or only the executed action, holding the skill block and teacher decisions fixed. A pairwise ranking loss over verified action outcomes is an orthogonal training signal, reported in Appendix~\ref{sec:appendix-alfworld-4b}, not a different deployment interface. Reward models in RLHF pipelines score candidate outputs or trajectories and can then guide policy optimization~\citep{christiano2017deep,ouyang2022training}. By contrast, a \method{} skill block does not score candidates at inference time; it directly conditions generation.

\paragraph{Continuous-prompt diagnostics and steering.}
Continuous prompts and activation-level steering methods raise the question of what remains inspectable when behavior is encoded in latent vectors~\citep{hambardzumyan2021warp,turner2023activation}. Later appendix probes use nearest-token and cosine-similarity diagnostics only to test how far a learned skill block moves from its text-derived initialization, not whether the skill block is semantically equivalent to the original skill. This supports a limited audit claim---the text skill remains useful for initialization, routing, and provenance---without treating the learned continuous skill block itself as human-interpretable.

\section{Extended Limitations}
\label{sec:appendix-limitations}

\paragraph{Serving and access.}
The deployment benefit of \method{} assumes a serving stack that can inject learned skill block embeddings. This is realistic for open or self-hosted models, but it makes the method less black-box than Markdown skill optimization. Training and serving require access to the target model's embedding interface, and the learned skill block may be tied to a specific tokenizer, prompt template, decoding setup, or inference harness. Deployment therefore couples the learned skill to a frozen-backbone serving path. A broader self-improvement system such as SIA~\citep{hebbar2026siaselfimprovingai} suggests a natural integration point: a feedback controller could choose among harness edits, parameter updates, and soft-skill block updates depending on whether the bottleneck appears to be scaffold engineering, domain knowledge, or compact input-side behavioral control.

\paragraph{Auditability.}
A learned skill block is harder to inspect than a natural-language skill. One practical compromise is to keep the Markdown skill as metadata for retrieval, provenance, and human audit, while injecting the learned skill block at inference time. In this design, the text skill remains useful, but it should not be treated as a complete explanation of the continuous behavior. Our experiments also use one skill per task and do not evaluate retrieval among many skills, so progressive-disclosure deployment remains an implication rather than a directly tested result.

\paragraph{Procedure versus formatting.}
The experiments do not fully distinguish robust procedure learning from answer-format learning. This matters because the strongest single-round results may partly come from inducing direct answers, short generations, or benchmark-specific extraction formats. Agentic ALFWorld makes the interface explicit: the fair 4B comparison forces the model to complete \actionopen{} rather than an open assistant turn. After that alignment, action-only supervision matches full-response next-token training in aggregate success, so we do not treat a loss on rationale tokens as the method. Masking those tokens while keeping gold rationale as teacher-forced context is a different---and harmful---condition. Evaluating the same trained skill block without retraining under instruction paraphrasing, formatting wrappers, and a fixed irrelevant-context block preserves a large trained-versus-initialization gap (Table~\ref{tab:alfworld-robustness}). That argues against an explanation based solely on the exact original prompt surface. Paired trajectory statistics on the clean interface further associate the gain with sequential action selection---fewer \nothinghappens{} replies and repetitive loops---rather than with improved recovery after an isolated rejection (Table~\ref{tab:alfworld-behavior-full}). Neither result establishes a general task policy, and other harness conventions---tool names, observation parsers, stopping rules, and answer extractors---remain untested. Appendix~\ref{sec:appendix-alfworld-mechanism} further localizes the action-only update: it is a small continuous offset of the text initialization, read in the earliest layers of this frozen 4B model. That localization uses one checkpoint. The 134-game path test confirms only the halfway interpolation and one same-norm random direction, and a 32-question SearchQA log-probability slice did not favor the trained skill, so the early-layer readout is not a cross-task claim.\xj{New limitation paragraph for the 2026-09-26 mechanism results. It belongs with the new appendix even if the main-text subsection is cut.}

\paragraph{Transfer and composition.}
Unlike Markdown skills, continuous skill blocks are not directly portable across models. A skill block trained for one embedding space may not transfer to another, even within the same model family. Vocabulary-projection transfer is one possible diagnostic: map each source skill block vector to a sparse distribution over source tokens, then reconstruct a target skill block using aligned token strings or shared vocabulary IDs. The measured transfer is task-dependent: DocVQA retains useful gains, whereas LiveMath requires target-side training in the evaluated settings.

The present experiments also train one soft skill per task. They do not show that soft skills compose, route, or transfer across related tasks. The ALFWorld concatenation diagnostic in Appendix~\ref{sec:appendix-composition} rules out a trivial $32{\to}64$ length confound, but it is not a factorial composition test: $A$ did not improve held-out $A$-only decisions, $B$ had no uncontaminated $B$-only holdout, and every $A{+}B$ evaluation episode also appears in component training. SearchQA is a poor substitute for that missing geometry. Semantically meaningful operations rarely co-occur as $A$-only / $B$-only / $A{+}B$; the pairs with enough counts are lexical templates rather than reusable procedures. Skill neologisms specifically target composition~\citep{berthon2026skillneologisms}; KV-Skill reports independently loadable operators through a shared interface~\citep{han2026kvskill}. A stronger version of \method{} would need independently learnable components, task-disjoint $A{+}B$ evaluation, and a composition operator that is not just sequence concatenation of independently optimized skill blocks. This is especially important for agentic settings, where useful behavior may be distributed across retrieval, planning, tool use, verification, and final-answer formatting.

\section{Experiment Setting}

The first appendix tables summarize the method taxonomy, data splits, and benchmark regimes used by the main experiments.

\begin{table*}[t]
\caption{Comparison of skill optimization methods and related adaptation baselines.}
\label{tab:baselines}
\centering
\small
\setlength{\tabcolsep}{4pt}
\renewcommand{\arraystretch}{1.12}
\begin{tabularx}{\textwidth}{@{}l l X c X@{}}
\toprule
Method
& Optimized object
& Training signal
& Rollouts?
& Inference-time deployment \\
\midrule
No skill
& None
& None
& No
& Task prompt only; frozen model \\

Manual hard skill
& Markdown skill
& Human writing/editing
& No
& Long text skill in context \\

SkillOpt hard skill
& Markdown skill
& Scored rollouts and validation feedback
& Yes
& Optimized text skill in context \\

Prompt optimization
& Prompt text
& Black-box scores, examples, or reflections
& Usually
& Optimized text prompt in context \\

LoRA / adapter
& Adapter weights
& Supervised trajectories or RL feedback
& Optional
& Adapter-aware serving; extra weights loaded \\

Prefix / P-tuning
& Virtual tokens
& Supervised trajectories or RL feedback
& Optional
& Frozen model plus learned continuous skill block \\

\method{}
& Text-initialized soft skill
& Gradient signal from answers or successful trajectories
& No
& Frozen model plus compact learned skill skill block \\

Hard skill + soft delta
& Text skill + skill block
& Answers, trajectories, or reward feedback
& Optional
& Text skill remains; skill block provides compact correction \\
\bottomrule
\end{tabularx}
\end{table*}

\begin{table}[htbp]
\caption{Dataset split sizes and experiment configuration. Main-text experiments summarize the total nominal train size and the smaller usable trajectory counts for the agentic tasks.}
\label{tab:data-splits}
\begin{center}
\begin{tabular}{lrrrrr}
\toprule
Task & Train & Val & Test & Total & Usable train \\
\midrule
DocVQA & 107 & 53 & 374 & 534 & 107 \\
SearchQA & 400 & 200 & 1{,}400 & 2{,}000 & 400 \\
LiveMath & 35 & 18 & 124 & 177  & 35 \\
ALFWorld & 39 & 18 & 134 & 191 & 31 \\
OfficeQA & 50 & 24 & 172 & 246 & 31\\
Spreadsheet & 80 & 40 & 280 & 400 & 61 \\
\bottomrule
\end{tabular}
\end{center}
\end{table}

\begin{table}[htbp]
\caption{Benchmark regimes and training supervision. Single-round QA uses answer NTP. OfficeQA and Spreadsheet stay on full-response trajectory NTP. ALFWorld additionally ablates the supervised token set on the same skill block.}
\label{tab:task-status}
\begin{center}
\resizebox{\linewidth}{!}{
\begin{tabular}{llll}
\toprule
Task & Regime & NTP target & Expected relation to SkillOpt \\
\midrule
SearchQA & Single-round QA & Ground-truth answer, no CoT & Answer/reasoning compression \\
LiveMath & Single-round QA & Ground-truth answer, no CoT & Answer/reasoning compression \\
DocVQA & Single-round QA & Ground-truth answer, no CoT & Answer/reasoning compression \\
OfficeQA & Agentic execution & Successful trajectory (SFT) & SkillOpt-artifact refinement \\
Spreadsheet & Agentic execution & Successful trajectory (SFT) & SkillOpt-artifact refinement \\
ALFWorld & Agentic execution & Full-response or action-only NTP; masked-rationale diagnostic & Same skill block; 4B forced-action vs 35B free-form \\
\bottomrule
\end{tabular}
}
\end{center}
\end{table}

\begin{table}[htbp]
\caption{Full single-round QA comparison on Qwen3.5--4B, including Human skill and TextGrad, which remain appendix-only relative to Table~\ref{tab:main-results}. \method{} reports mean $\pm$ standard deviation over three seeds under the default placement. Deltas are versus no skill.}
\label{tab:main-results-full}
\begin{center}
\resizebox{\linewidth}{!}{
\begin{tabular}{lccc}
\toprule
Method & SearchQA & LiveMath & DocVQA \\
\midrule
No skill & \acc{68.1}{+0.0} & \acc{22.4}{+0.0} & \acc{86.9}{+0.0} \\
Human skill & \acc{66.3}{-1.8} & \acc{18.4}{-4.0} & \acc{87.8}{+0.9} \\
LLM skill & \acc{65.0}{-3.1} & \acc{28.8}{+6.4} & \acc{88.0}{+1.1} \\
Trace2Skill~\citep{ni2026trace2skilldistilltrajectorylocallessons} & \acc{68.5}{+0.4} & \acc{27.2}{+4.8} & \acc{88.0}{+1.1} \\
TextGrad~\citep{yuksekgonul2024textgrad} & \acc{60.7}{-7.4} & \acc{10.6}{-11.8} & \acc{85.6}{-1.3} \\
GEPA~\citep{agrawal2025gepa} & \acc{68.6}{+0.5} & \acc{28.8}{+6.4} & \acc{85.1}{-1.8} \\
SkillOpt~\citep{yang2026skillopt} & \acc{71.2}{+3.1} & \acc{52.0}{+29.6} & \acc{89.0}{+2.1} \\
\midrule
LoRA~\citep{hu2021lora} & \best{\acc{78.6}{+10.5}} & \acc{58.9}{+36.5} & \acc{87.7}{+0.8} \\
\method{} & \acc{$75.7 \pm 1.9$}{+7.6} & \best{\acc{$64.5 \pm 2.4$}{+42.1}} & \acc{$88.2 \pm 1.4$}{+1.3} \\
\bottomrule
\end{tabular}
}
\end{center}
\end{table}

\section{Additional Ablations}

Placement appears in the main text (Table~\ref{tab:placement}). This section reports the full skill-length, data-scaling, model-scale, transfer, checkpoint, and geometry diagnostics.

\begin{table}[ht]
\caption{Skill-length sweep with Qwen3.5--4B, reported as seed-1 test-hard accuracy. Auto lengths are listed in task-column order; DocVQA uses 105 tokens for the NL-initialized \texttt{skill\_section} run and 407 for the existing \texttt{prompt\_start} run.}
\label{tab:prefix-length-hard}
\begin{center}
\resizebox{0.95\linewidth}{!}{
\begin{tabular}{llccc}
\toprule
Placement & Skill length & SearchQA & LiveMath & DocVQA \\
\midrule
\texttt{No Skill} & \tbd & \acc{68.1}{+0.0} & \acc{22.4}{+0.0} & \acc{86.9}{+0.0} \\
\texttt{SkillOpt} & \tbd & \acc{71.2}{+3.1} & \acc{52.0}{+29.6} & \acc{89.0}{+2.1} \\
\midrule
\texttt{prompt\_start} & 8 & \acc{75.6}{+7.5} & \acc{56.5}{+34.1} & \acc{87.4}{+0.5} \\
\texttt{skill\_section} & 8 & \acc{75.9}{+7.8} & \acc{43.6}{+21.2} & \acc{88.8}{+1.9} \\
\texttt{prompt\_start} & 32 & \acc{76.4}{+8.3} & \acc{60.5}{+38.1} & \acc{89.3}{+2.4} \\
\texttt{skill\_section} & 32 & \best{\acc{77.3}{+9.2}} & \best{\acc{66.9}{+44.5}} & \best{\acc{89.8}{+2.9}} \\
\texttt{prompt\_start} & 256 & \acc{76.9}{+8.8} & \acc{58.1}{+35.7} & \acc{89.3}{+2.4} \\
\texttt{skill\_section} & 256 & \acc{76.1}{+8.0} & \acc{63.7}{+41.3} & \acc{88.2}{+1.3} \\
\texttt{prompt\_start} & auto (2035/671/407) & \acc{75.6}{+7.5} & \acc{55.6}{+33.2} & \acc{89.3}{+2.4} \\
\texttt{skill\_section} & auto (2035/671/105) & \acc{75.6}{+7.5} & \acc{59.7}{+37.3} & \acc{88.2}{+1.3} \\
\bottomrule
\end{tabular}
}
\end{center}
\end{table}

\paragraph{Training-data scaling.}
Exact values for the main-text scaling curves (Figure~\ref{fig:data-scaling}) are listed below.

\begin{table}[ht]
\caption{Train-data scaling study on Qwen3.5--4B. Deltas are versus no skill. Full-data NL-skill values use three-seed means except SearchQA at \texttt{skill\_section} (seed 1); no seed dispersion is reported for the reduced-data runs. DocVQA \texttt{skill\_section} uses seed 1 at 10\% and 50\% (11 and 54 examples); its full-data seed-1 score is 89.8 on 107 training examples (Table~\ref{tab:docvqa-skill-section-runs}).}
\label{tab:data-scaling-placeholder}
\begin{center}
\small
\resizebox{\linewidth}{!}{
\begin{tabular}{llccc}
\toprule
Task & Training fraction & LoRA & \method{} \texttt{prompt\_start} & \method{} \texttt{skill\_section} \\
\midrule
SearchQA & 10\% & \acc{72.4}{+4.3} & \acc{71.1}{+3.0} & \acc{70.2}{+2.1} \\
SearchQA & 50\% & \acc{77.4}{+9.3} & \acc{74.0}{+5.9} & \acc{72.6}{+4.5}\\
SearchQA & 100\% & \best{\acc{78.6}{+10.5}} & \acc{$76.4$}{+8.3} & \acc{$77.3$}{+9.2}\\
\addlinespace
LiveMath & 10\% & \acc{33.1}{+10.7} & \acc{37.1}{+14.7} & \acc{30.6}{+8.2} \\
LiveMath & 50\% & \acc{62.1}{+39.7} & \acc{61.3}{+38.9} & \acc{64.5}{+42.1}\\
LiveMath & 100\% & \acc{58.9}{+36.5} & \acc{$50.0$}{+27.6} & \best{\acc{$64.5$}{+42.1}}\\
\addlinespace
DocVQA & 10\% & \acc{86.9}{+0.0} & \acc{88.0}{+1.1} & \acc{83.7}{-3.2} \\
DocVQA & 50\% & \acc{88.8}{+1.9} & \acc{89.6}{+2.7} & \acc{88.2}{+1.3} \\
DocVQA & 100\% & \acc{87.7}{+0.8} & \acc{$88.2$}{+1.3} & \best{\acc{$89.9$}{+3.0}}\\
\bottomrule
\end{tabular}
}
\end{center}
\end{table}

\paragraph{Model scale.}
Table~\ref{tab:scale-transfer} summarizes model scale and direct transfer; Table~\ref{tab:model-size-hard} reports all length-32 placements. Scaling improves SearchQA and DocVQA, while LiveMath is non-monotonic.

\begin{table}[ht]
\caption{Model scale and direct cross-model skill-block transfer, seed-1 test-hard accuracy. Trained rows retrain a length-32 block on the named target. Transfer rows project a source block through the vocabulary and evaluate without target-side NTP.}
\label{tab:scale-transfer}
\begin{center}
\small
\begin{tabular}{llccc}
\toprule
Setting & Model / direction & SearchQA & LiveMath & DocVQA \\
\midrule
Trained & Qwen3.5--4B & 77.3 & 66.9 & 89.3 \\
Trained & Qwen3.5--9B & 80.9 & 66.9 & 92.5 \\
Trained & Qwen3.6--35B-A3B & 82.3 & 50.0 & 93.3 \\
No skill & Qwen3.5--4B & 68.1 & 22.4 & 86.9 \\
No skill & Qwen3.6--35B-A3B & 72.7 & 31.2 & 87.6 \\
\midrule
Transfer & 4B $\rightarrow$ 35B & 74.9 & 15.3 & 93.0 \\
Transfer & 35B $\rightarrow$ 4B & 68.3 & 14.5 & 90.4 \\
\bottomrule
\end{tabular}
\end{center}
\end{table}

\begin{table}[htbp]
\caption{Model-size sweep for length-32 \method{}, reported as seed-1 test-hard accuracy.}
\label{tab:model-size-hard}
\begin{center}
\resizebox{0.95\linewidth}{!}{
\begin{tabular}{llccc}
\toprule
Model & Placement & SearchQA & LiveMath & DocVQA \\
\midrule
Qwen3.5--4B & No Skill & \acc{68.1}{+0.0} & \acc{22.4}{+0.0} & \acc{86.9}{+0.0} \\
Qwen3.5--4B & SkillOpt & \acc{71.2}{+3.1} & \acc{52.0}{+29.6} & \acc{89.0}{+2.1} \\
Qwen3.5--4B & \texttt{prompt\_start} & \acc{76.4}{+8.3} & \acc{60.5}{+38.1} & \acc{89.3}{+2.4} \\
Qwen3.5--4B & \texttt{skill\_section} & \acc{77.3}{+9.2} & \acc{66.9}{+44.5} & \acc{89.8}{+2.9} \\
\midrule
Qwen3.5--9B & \texttt{prompt\_start} & 80.9 & \best{66.9} & 92.5 \\
Qwen3.5--9B & \texttt{skill\_section} & 78.1 & 64.5 & \tbd{} \\
\midrule
Qwen3.6--35B-A3B & No Skill & \acc{72.7}{+0.0} & \acc{31.2}{+0.0} & \acc{87.6}{+0.0} \\
Qwen3.6--35B-A3B & SkillOpt & \acc{80.3}{+7.6} & \acc{41.6}{+10.4} & \acc{91.4}{+3.8} \\
Qwen3.6--35B-A3B & \texttt{prompt\_start} & \best{\acc{82.3}{+9.6}} & \acc{50.0}{+18.8} & \best{\acc{93.3}{+5.7}} \\
Qwen3.6--35B-A3B & \texttt{skill\_section} & \acc{82.1}{+9.4} & \acc{45.2}{+14.0} & \tbd{} \\
\bottomrule
\end{tabular}
}
\end{center}
\end{table}

\paragraph{Cross-model transfer.}
The model-size sweep retrains a separate skill block for each target model. Table~\ref{tab:cross-model-transfer} is the full transfer diagnostic summarized in Table~\ref{tab:scale-transfer}: each source-model skill block is projected through the vocabulary and reconstructed in the target embedding space, then evaluated without target-side NTP. Transfer is task-dependent rather than uniformly portable. DocVQA transfers well in both directions. SearchQA transfers only partially. LiveMath is the clearest failure case, falling below no skill in both directions. The learned skill block can contain transferable task bias for some answer-extraction settings, but it remains tightly coupled to model-specific embedding geometry.

\begin{table}[htbp]
\caption{Direct cross-model skill block transfer without target-side training. Transfer rows report seed-1 test-hard and test-soft accuracy. Target-trained \method{} uses the corresponding target-model seed-1 test-hard result.}
\label{tab:cross-model-transfer}
\begin{center}
\resizebox{\linewidth}{!}{
\begin{tabular}{llcccc}
\toprule
Task & Transfer direction & Target no skill & Transfer hard & Transfer soft & Target-trained \method{} \\
\midrule
SearchQA & Qwen3.5--4B $\rightarrow$ Qwen3.6--35B-A3B & \acc{72.7}{+0.0} & \acc{74.9}{+2.2} & 84.2 & \acc{82.1}{+9.4} \\
SearchQA & Qwen3.6--35B-A3B $\rightarrow$ Qwen3.5--4B & \acc{68.1}{+0.0} & \acc{68.3}{+0.2} & 76.3 & \acc{77.3}{+9.2} \\
LiveMath & Qwen3.5--4B $\rightarrow$ Qwen3.6--35B-A3B & \acc{31.2}{+0.0} & \acc{15.3}{-15.9} & 15.3 & \acc{45.2}{+14.0} \\
LiveMath & Qwen3.6--35B-A3B $\rightarrow$ Qwen3.5--4B & \acc{22.4}{+0.0} & \acc{14.5}{-7.9} & 14.5 & \acc{66.9}{+44.5} \\
DocVQA & Qwen3.5--4B $\rightarrow$ Qwen3.6--35B-A3B & \acc{87.6}{+0.0} & \acc{93.0}{+5.4} & 95.8 & \acc{93.3}{+5.7} \\
DocVQA & Qwen3.6--35B-A3B $\rightarrow$ Qwen3.5--4B & \acc{86.9}{+0.0} & \acc{90.4}{+3.5} & 94.4 & \acc{89.3}{+2.4} \\
\bottomrule
\end{tabular}
}
\end{center}
\end{table}

\begin{figure}[htbp]
\begin{center}
\includegraphics[width=0.49\linewidth]{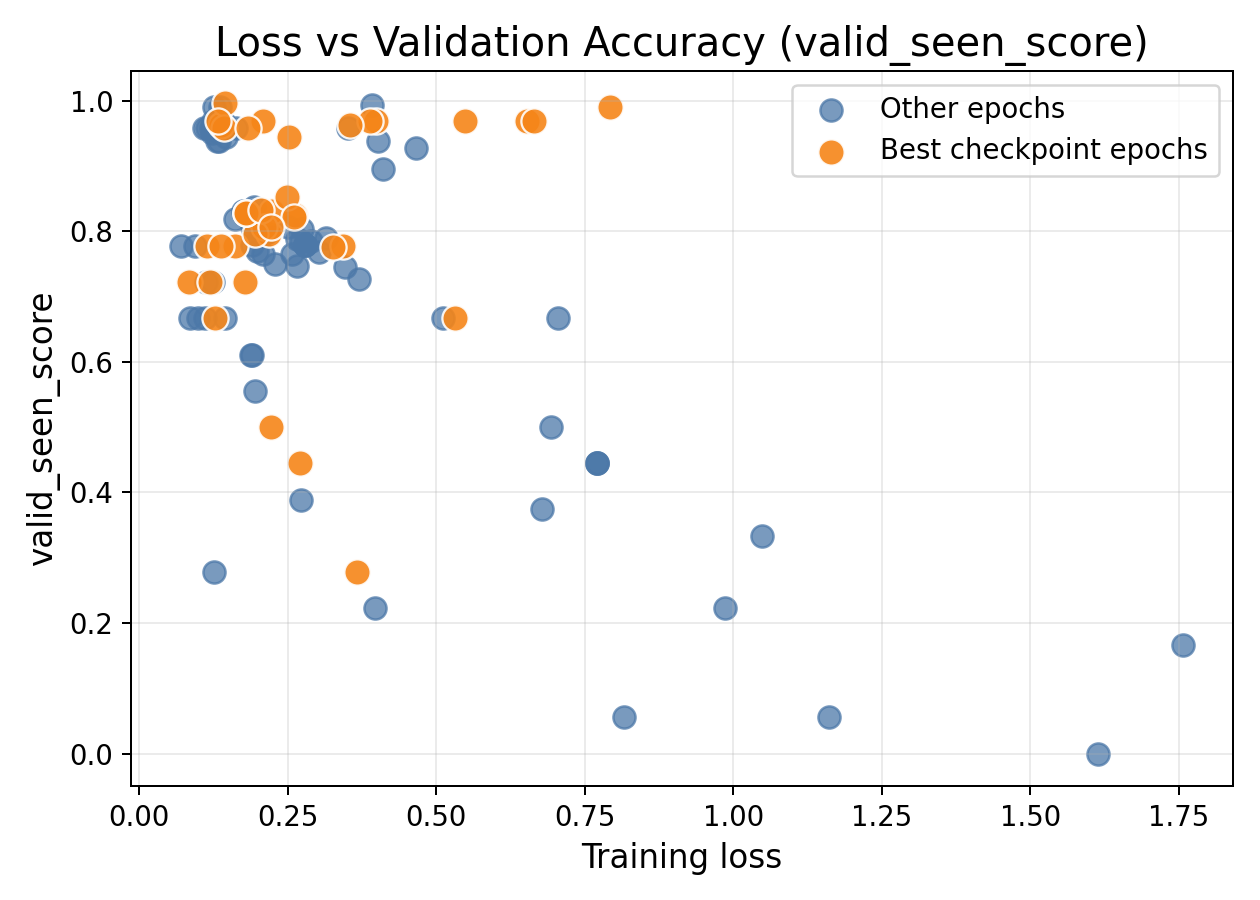}
\hfill
\includegraphics[width=0.49\linewidth]{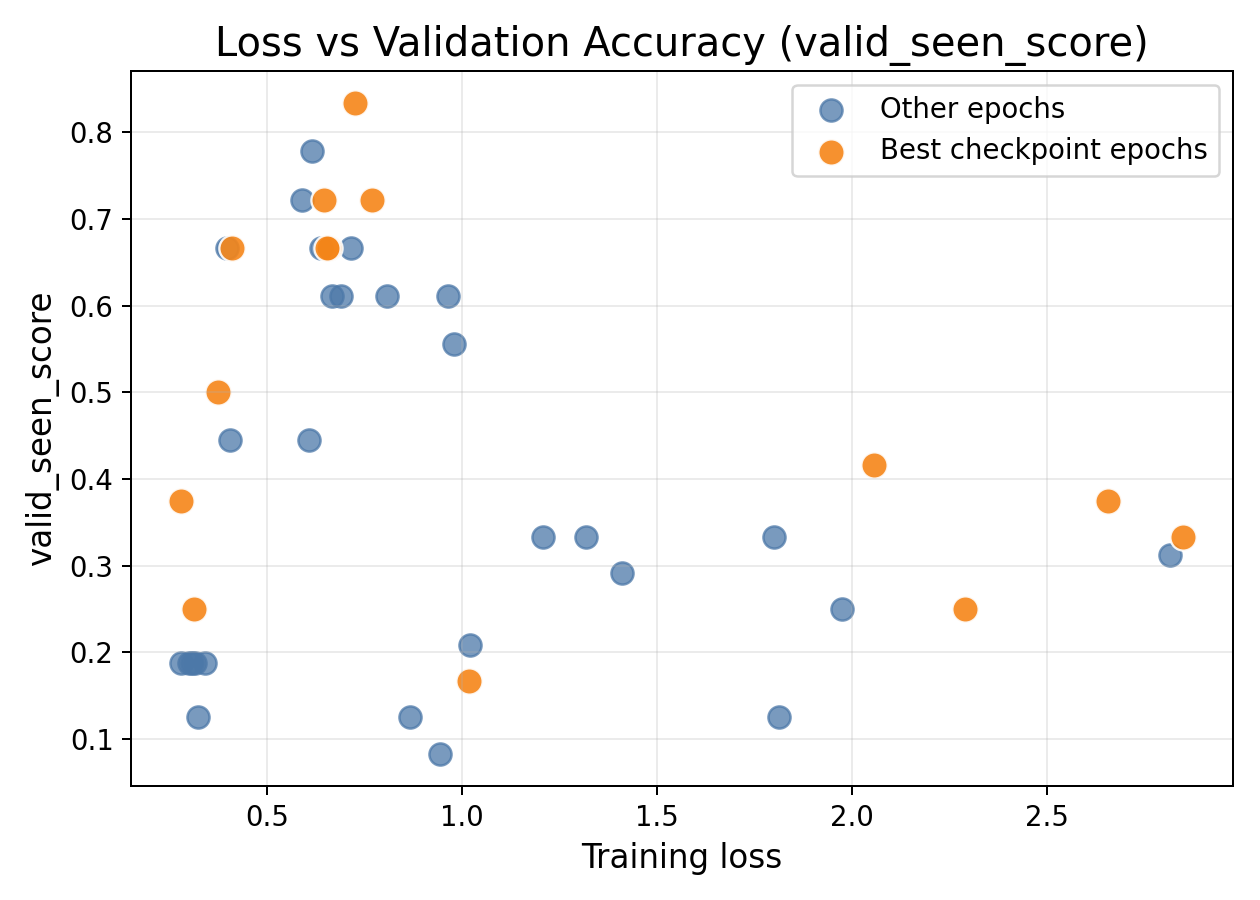}
\end{center}
\caption{Training loss versus held-out validation task accuracy for single-round runs (left) and agentic runs (right). Orange points mark the checkpoint selected by validation task accuracy within each run; blue points are other checkpoints. Lower loss is not sufficient to identify the best task checkpoint, especially in the agentic setting.}
\label{fig:loss-vs-validation-accuracy}
\end{figure}

\paragraph{Loss versus task success.}
Figure~\ref{fig:loss-vs-validation-accuracy} compares NTP loss with held-out task accuracy across checkpoints. Very high loss usually corresponds to poor behavior, but among lower-loss checkpoints the correlation with task success is weak. The validation-selected checkpoint is often not the minimum-loss checkpoint. In agentic runs this issue is more visible: masked-rationale training drives action NLL near zero while held-out success falls.

\begin{figure}[htbp]
\begin{center}
\includegraphics[width=0.55\linewidth]{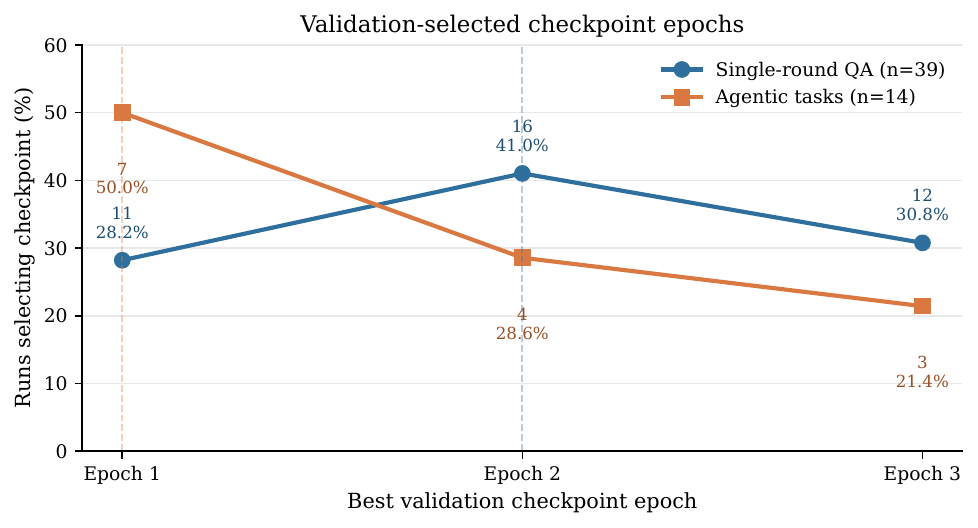}
\end{center}
\caption{Validation-selected checkpoint epochs for \method{} training runs. Single-round runs peak across all three epochs; agentic runs more often peak at epoch 1.}
\label{fig:validation-checkpoint-frequency}
\end{figure}

\paragraph{Validation-selected checkpoints.}
Validation often selects a checkpoint before the final epoch. Figure~\ref{fig:validation-checkpoint-frequency} summarizes this effect. In single-round runs, the best checkpoint occurs at epoch 1, 2, and 3 in 30.8\%, 30.8\%, and 38.5\% of runs, respectively. In agentic runs, selection is more front-loaded: 50.0\% of runs peak at epoch 1. Prefix tuning can overfit even though only a small number of virtual tokens is trained.

\paragraph{Skill geometry.}
We use skill block-geometry diagnostics to ask whether different training runs converge to similar learned conditioning states. For each learned skill block vector $p_i \in \mathbb{R}^d$ and vocabulary embedding $e_j \in \mathbb{R}^d$, we compute
\begin{equation}
    \cos(p_i, e_j) = \frac{p_i^\top e_j}{\lVert p_i \rVert_2 \lVert e_j \rVert_2}.
\end{equation}
Nearest-token probes rank vocabulary items by this score for each skill block position. For text-initialized skill blocks, we also compute $\cos(p_i, p_i^{\mathrm{init}})$, averaged over skill block positions. These probes are diagnostic only: they can reveal geometric drift or clustering, but they do not prove that the continuous skill block preserves the semantics of the initializing skill.

\begin{figure}[htbp]
\centering
\includegraphics[width=0.33\linewidth]{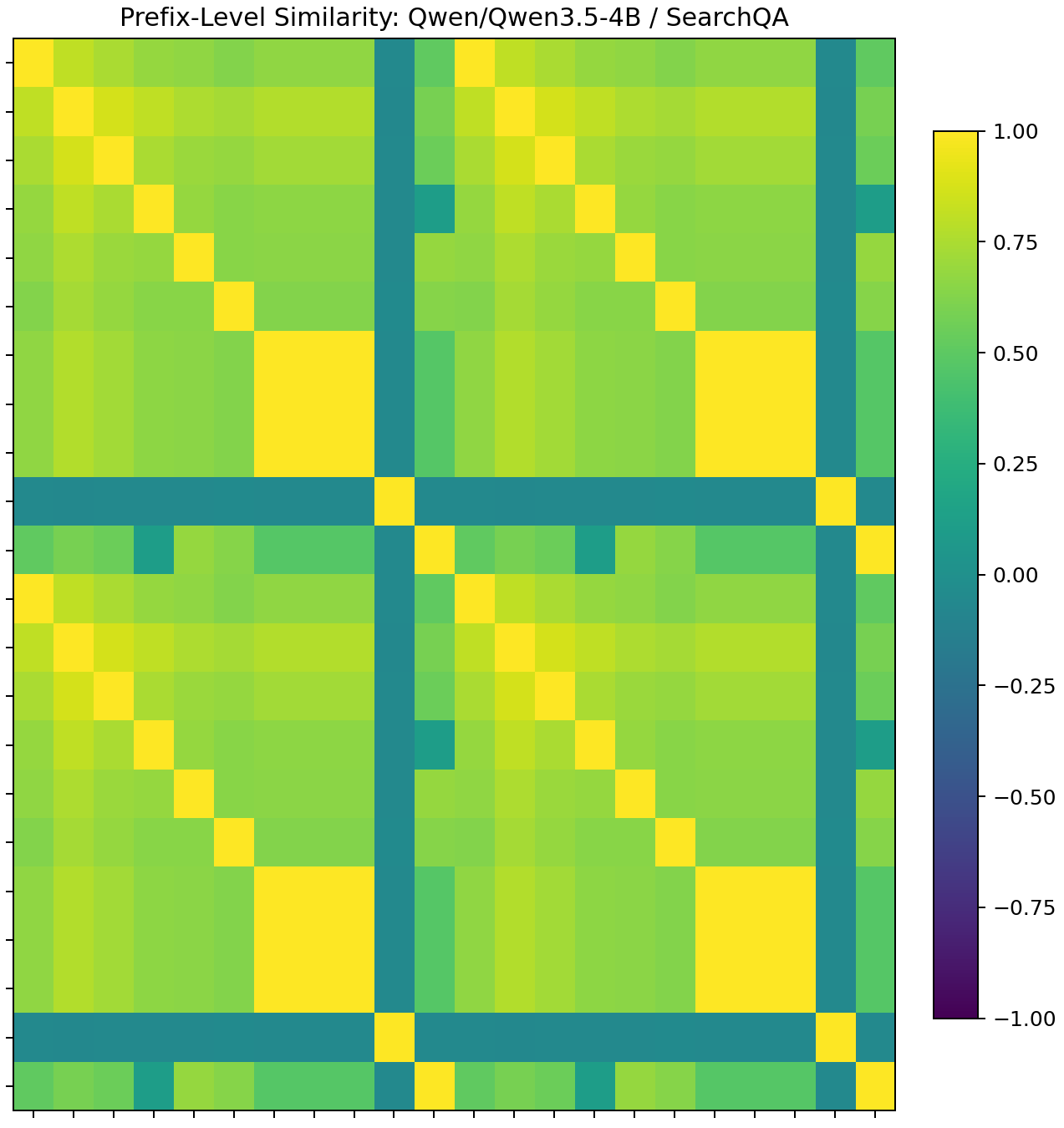}
\hfill
\includegraphics[width=0.32\linewidth]{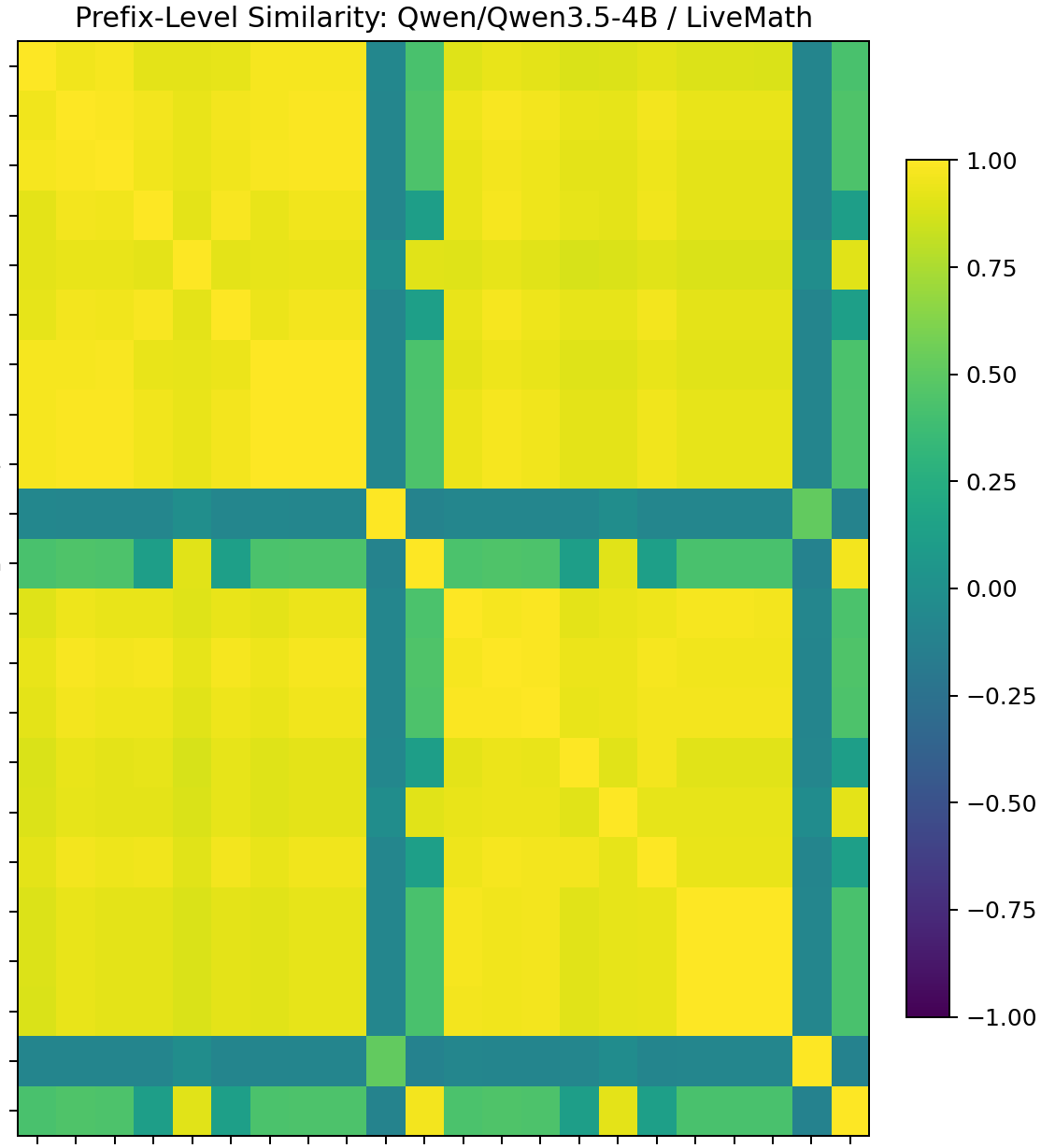}
\hfill
\includegraphics[width=0.33\linewidth]{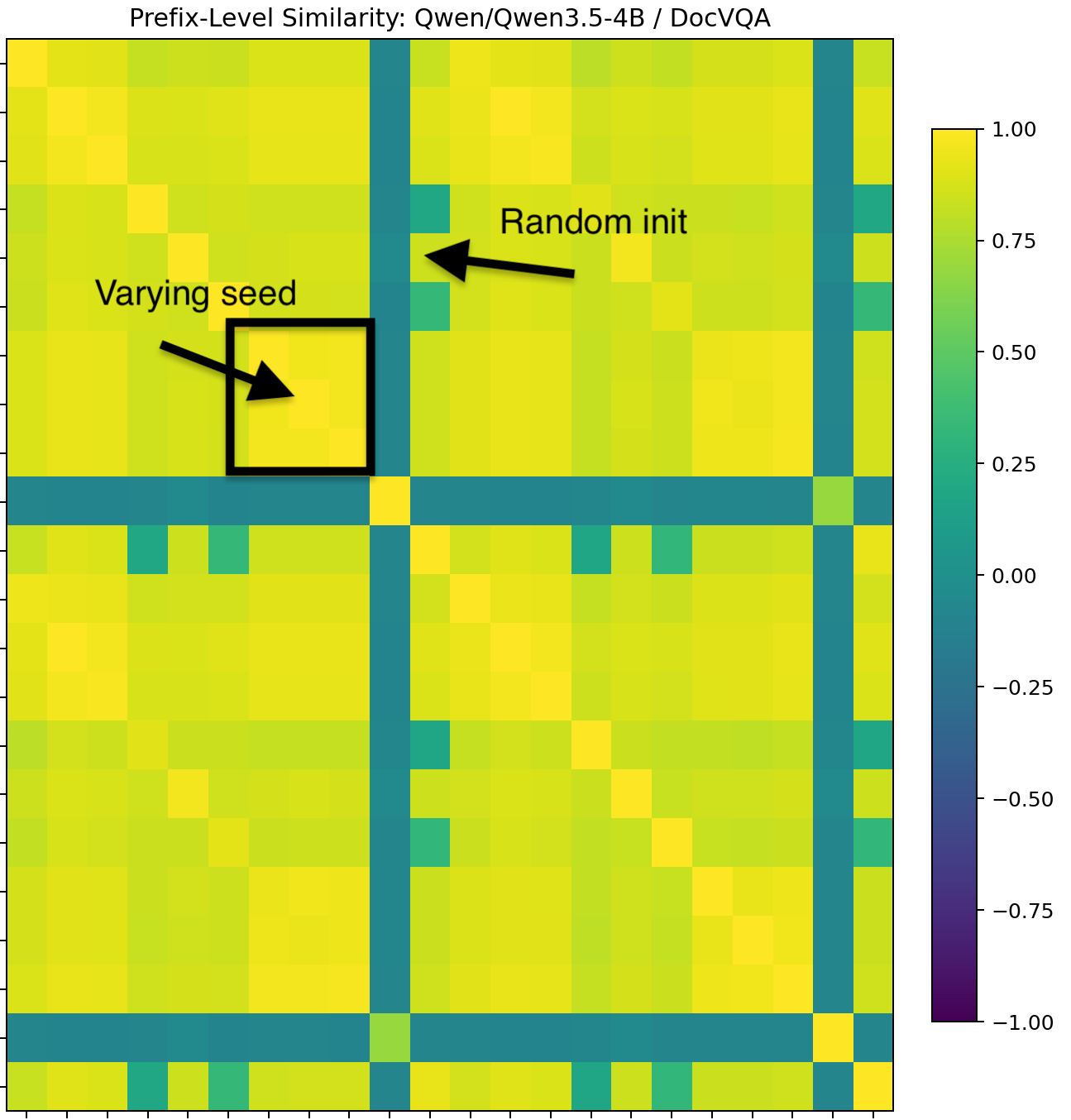}
\caption{Skill-level cosine-similarity heatmaps for Qwen3.5--4B on SearchQA (left), LiveMath (middle), and DocVQA (right). Dark stripes correspond to random or mean-pooled initialized skill blocks with lower similarity to the rest. The heatmaps are diagnostics of run-to-run geometry, not evidence that the skill block preserves the semantics of the original skill text.}
\label{fig:prefix-similarity-4b}
\end{figure}

\section{Additional Results}
\label{sec:appendix-additional-results}

The reproduced SkillOpt baselines use the public artifacts and the available evaluation harness, but they are not exact reproductions of the paper's OfficeQA environment. The authors clarified that their OfficeQA experiments used a Google-backed web-search tool in which candidate files served as hints rather than as the full accessible evidence set. By contrast, the public offline mode uses local document tools, and the public custom-search path requires a compatible authenticated search service. The public release also does not fully pin Qwen inference parameters, so differences in stopping criteria, decoding budget, and serving configuration may change reproduced Qwen results.

In the auxiliary QA tables, ``hard'' and ``soft'' refer to the evaluator used to score a generated answer, not to whether the inference input contains a hard or soft skill. ``Hard'' accuracy requires the exact task-level criterion used in the main results. ``Soft'' accuracy uses the more permissive answer-equivalence evaluator used for auxiliary diagnostics. Validation rows score the held-out selection split; test rows score the final held-out test split.

\subsection{DocVQA Placement in the Reported Experiments}
\label{sec:appendix-docvqa-placement}

The headline DocVQA result remains the three-seed \texttt{prompt\_start} result. Earlier runs labeled \texttt{skill\_section} also prepended the embeddings in the vision-language path; those runs are excluded from the placement comparison. The new \texttt{skill\_section} runs use Qwen3.5--4B and insertion at the skill-section slot, logged at token index 14. The main setting uses natural-language initialization and 107 training examples; the seed-1 sweeps vary length, training fraction, or initialization. We select the epoch with the highest validation soft score and score its checkpoint once on the 374-example test set. Epoch 1 is selected for the length-32 main runs and length-8 run; epochs 2 and 3 do not improve their validation soft scores. The length-256 and 10\%-data runs select epoch 2; the other completed sweeps select epoch 1.

At length 32, seeds 1--3 reach $89.9\pm0.4$ test hard and $94.1\pm0.2$ test soft, with validation hard and soft scores of $94.8\pm1.8$ and $96.2\pm0.6$. The test-hard mean is 1.7 points above the existing \texttt{prompt\_start} mean of $88.2\pm1.4$ (Table~\ref{tab:placement}). The earlier seed-42 audit is excluded from this aggregate. At length 8, seed 1 reaches 88.8 test hard and 93.3 test soft, compared with 87.4 test hard for \texttt{prompt\_start}. Tables~\ref{tab:docvqa-skill-section-runs} and~\ref{tab:docvqa-skill-section-sweeps} give all ten completed runs. DocVQA skill-section runs on 9B and 35B have not been run.

The additional seed-1 sweeps show that the placement gain depends on configuration. At length 256, \texttt{skill\_section} scores 88.2 test hard versus 89.3 for \texttt{prompt\_start}. The auto-length runs score 88.2 versus 89.3, but use different lengths (105 versus 407 tokens). At 10\% and 50\% data, skill-section scores 83.7 and 88.2 versus prompt-start's 88.0 and 89.6. The 10\%-data run's epoch-1 validation score is zero, so that checkpoint is rejected; the selected epoch-2 checkpoint generates 14.0 tokens on average. Mean-pooled initialization matches prompt-start at 89.6 test hard. SkillOpt-artifact initialization reaches 89.0 versus 88.5 for prompt-start, using a 407-token skill block. These initialization results compare different skill lengths: NL and mean-pooled use 32 tokens.

\begin{table}[ht]
\caption{DocVQA \texttt{skill\_section} main runs and length-8 run on Qwen3.5--4B with natural-language initialization and 107 training examples. Every selected checkpoint is from epoch 1. Hard and soft scores are percentages; output length is the mean generated-token count on test.}
\label{tab:docvqa-skill-section-runs}
\centering
\small
\begin{tabular}{rrrrrrr}
\toprule
Length & Seed & Val hard & Val soft & Test hard & Test soft & Output tokens \\
\midrule
32 & 1 & 93.8 & 95.8 & 89.8 (336/374) & 94.3 & 11.1 \\
32 & 2 & 93.8 & 95.8 & 90.4 (338/374) & 94.1 & 11.3 \\
32 & 3 & 96.9 & 96.9 & 89.6 (335/374) & 93.9 & 11.1 \\
8 & 1 & 96.9 & 96.9 & 88.8 (332/374) & 93.3 & 12.1 \\
\bottomrule
\end{tabular}
\end{table}

\begin{table}[ht]
\caption{Additional DocVQA \texttt{skill\_section} sweeps on Qwen3.5--4B, seed 1. Scores are hard / soft percentages. All runs use insertion at token index 14, validation-selected checkpoints, and 374 test examples. Auto length uses all 105 NL-skill tokens; the SkillOpt artifact uses 407 tokens.}
\label{tab:docvqa-skill-section-sweeps}
\centering
\small
\begin{tabular}{lrrrll}
\toprule
Setting & Length & Train & Epoch & Val hard / soft & Test hard / soft \\
\midrule
NL skill & 256 & 107 & 2 & 93.8 / 95.8 & 88.2 / 93.5 \\
NL skill, auto & 105 & 107 & 1 & 93.8 / 98.6 & 88.2 / 92.3 \\
NL skill, 10\% data & 32 & 11 & 2 & 87.5 / 95.5 & 83.7 / 89.3 \\
NL skill, 50\% data & 32 & 54 & 1 & 96.9 / 96.9 & 88.2 / 93.5 \\
Mean-pooled & 32 & 107 & 1 & 93.8 / 95.8 & 89.6 / 93.9 \\
SkillOpt artifact & 407 & 107 & 1 & 96.9 / 96.9 & 89.0 / 93.5 \\
\bottomrule
\end{tabular}
\end{table}

\begin{table}[h]
\caption{Compression diagnostics for the single-round QA setting. \method{} results are from the main seed-1 Qwen3.5--4B run; SkillOpt (reproduced) rows are from our inference run using the released final SkillOpt artifacts with the available Qwen3.5--4B harness. Reported SkillOpt accuracies are the Qwen3.5--4B numbers from \citet{yang2026skillopt}.}
\label{tab:appendix-compression-diagnostics}
\begin{center}
\resizebox{\linewidth}{!}{
\begin{tabular}{llrrrr}
\toprule
Task & Method & Skill tokens & Average output tokens & Reproduced accuracy & Reported SkillOpt accuracy  \\
\midrule
SearchQA & SkillOpt (reproduced) & 2035 & 98.4 & 72.0 & 71.2 \\
SearchQA & \method{} & 32 & 8.67 & 76.4 & -- \\
LiveMath & SkillOpt (reproduced) & 671 & 4422 & 37.1 & 52.0 \\
LiveMath & \method{} & 32 & 5.99 & 60.5 & -- \\
DocVQA & SkillOpt (reproduced) & 407  & 29.6 & 90.1 & 89.0 \\
DocVQA & \method{}  & 32 & 11.1 & 89.3 & -- \\
\bottomrule
\end{tabular}
}
\end{center}
\end{table}

\begin{table}[h]
\caption{Auxiliary main-setting aggregate metrics for Qwen3.5--4B, length 32. Each cell reports mean $\pm$ standard deviation over three seeds.}
\label{tab:appendix-main-aux}
\begin{center}
\resizebox{0.9\linewidth}{!}{
\begin{tabular}{llccc}
\toprule
Metric & Placement & SearchQA & LiveMath & DocVQA \\
\midrule
Test hard & \texttt{prompt\_start} & $76.4 \pm 0.0$ & $50.0 \pm 9.1$ & $88.2 \pm 1.4$ \\
Test hard & \texttt{skill\_section} & $75.7 \pm 1.9$ & $64.5 \pm 2.4$ & $89.9 \pm 0.4$ \\
Test soft & \texttt{prompt\_start} & $84.1 \pm 0.0$ & $50.0 \pm 9.1$ & $93.1 \pm 0.6$ \\
Test soft & \texttt{skill\_section} & $83.5 \pm 1.6$ & $64.5 \pm 2.4$ & $94.1 \pm 0.2$ \\
Validation hard & \texttt{prompt\_start} & $78.1 \pm 0.0$ & $63.0 \pm 3.2$ & $91.7 \pm 3.6$ \\
Validation hard & \texttt{skill\_section} & $77.6 \pm 1.8$ & $72.2 \pm 5.6$ & $94.8 \pm 1.8$ \\
Validation soft & \texttt{prompt\_start} & $82.8 \pm 0.0$ & $63.0 \pm 3.2$ & $95.7 \pm 0.2$ \\
Validation soft & \texttt{skill\_section} & $82.2 \pm 2.4$ & $72.2 \pm 5.6$ & $96.2 \pm 0.6$ \\
\bottomrule
\end{tabular}
}
\end{center}
\end{table}

\begin{table}[h]
\caption{Per-seed test-hard accuracy for the Qwen3.5--4B length-32 main setting.}
\label{tab:appendix-per-seed-hard}
\begin{center}
\begin{tabular}{llccc}
\toprule
Placement & Seed & SearchQA & LiveMath & DocVQA \\
\midrule
\texttt{prompt\_start} & 1 & 76.4 & 60.5 & 89.3 \\
\texttt{prompt\_start} & 2 & 76.4 & 45.2 & 88.8 \\
\texttt{prompt\_start} & 3 & 76.4 & 44.4 & 86.6 \\
\texttt{skill\_section} & 1 & 77.3 & 66.9 & 89.8 \\
\texttt{skill\_section} & 2 & 76.3 & 62.1 & 90.4 \\
\texttt{skill\_section} & 3 & 73.6 & 64.5 & 89.6 \\
\bottomrule
\end{tabular}
\end{center}
\end{table}

\begin{table}[h]
\caption{Auxiliary skill block-length metrics for Qwen3.5--4B, reported as seed-1 accuracy. DocVQA auto-length metrics are in Table~\ref{tab:docvqa-skill-section-sweeps}.}
\label{tab:appendix-length-aux}
\begin{center}
\resizebox{\linewidth}{!}{
\begin{tabular}{lllccc}
\toprule
Metric & Placement & Skill length & SearchQA & LiveMath & DocVQA \\
\midrule
Test soft & \texttt{prompt\_start} & 8 & 83.6 & 56.5 & 93.1 \\
Test soft & \texttt{skill\_section} & 8 & 83.6 & 43.6 & 93.3 \\
Test soft & \texttt{prompt\_start} & 32 & 84.1 & 60.5 & 93.6 \\
Test soft & \texttt{skill\_section} & 32 & 84.9 & 66.9 & 94.3 \\
Test soft & \texttt{prompt\_start} & 256 & 84.7 & 58.1 & 93.6 \\
Test soft & \texttt{skill\_section} & 256 & 83.2 & 63.7 & 93.5 \\
Validation hard & \texttt{prompt\_start} & 8 & 79.7 & 66.7 & 96.9 \\
Validation hard & \texttt{skill\_section} & 8 & 73.4 & 50.0 & 96.9 \\
Validation hard & \texttt{prompt\_start} & 32 & 78.1 & 66.7 & 93.8 \\
Validation hard & \texttt{skill\_section} & 32 & 79.7 & 77.8 & 93.8 \\
Validation hard & \texttt{prompt\_start} & 256 & 76.6 & 61.1 & 96.9 \\
Validation hard & \texttt{skill\_section} & 256 & 75.0 & 72.2 & 93.8 \\
Validation soft & \texttt{prompt\_start} & 8 & 85.3 & 66.7 & 96.9 \\
Validation soft & \texttt{skill\_section} & 8 & 81.0 & 50.0 & 96.9 \\
Validation soft & \texttt{prompt\_start} & 32 & 82.8 & 66.7 & 95.8 \\
Validation soft & \texttt{skill\_section} & 32 & 84.9 & 77.8 & 95.8 \\
Validation soft & \texttt{prompt\_start} & 256 & 80.6 & 61.1 & 96.9 \\
Validation soft & \texttt{skill\_section} & 256 & 79.6 & 72.2 & 95.8 \\
\bottomrule
\end{tabular}
}
\end{center}
\end{table}

\begin{table}[h]
\caption{Auxiliary model-sweep metrics for length-32 \method{}-NTP, reported as seed-1 accuracy. DocVQA \texttt{skill\_section} runs on 9B and 35B have not been started.}
\label{tab:appendix-model-aux}
\begin{center}
\resizebox{\linewidth}{!}{
\begin{tabular}{llccc}
\toprule
Metric / placement & Model & SearchQA & LiveMath & DocVQA \\
\midrule
Test soft, \texttt{prompt\_start} & Qwen3.5--4B & 84.1 & 60.5 & 93.6 \\
Test soft, \texttt{skill\_section} & Qwen3.5--4B & 84.9 & 66.9 & 94.3 \\
Test soft, \texttt{prompt\_start} & Qwen3.5--9B & 87.5 & 66.9 & 96.0 \\
Test soft, \texttt{skill\_section} & Qwen3.5--9B & 85.4 & 64.5 & \tbd{} \\
Test soft, \texttt{prompt\_start} & Qwen3.6--35B-A3B & 89.5 & 50.0 & 95.8 \\
Test soft, \texttt{skill\_section} & Qwen3.6--35B-A3B & 88.8 & 45.2 & \tbd{} \\
Validation hard, \texttt{prompt\_start} & Qwen3.5--4B & 78.1 & 66.7 & 93.8 \\
Validation hard, \texttt{skill\_section} & Qwen3.5--4B & 79.7 & 77.8 & 93.8 \\
Validation hard, \texttt{prompt\_start} & Qwen3.5--9B & 79.7 & 77.8 & 100.0 \\
Validation hard, \texttt{skill\_section} & Qwen3.5--9B & 78.1 & 77.8 & \tbd{} \\
Validation hard, \texttt{prompt\_start} & Qwen3.6--35B-A3B & 82.8 & 94.4 & 96.9 \\
Validation hard, \texttt{skill\_section} & Qwen3.6--35B-A3B & 84.4 & 94.4 & \tbd{} \\
Validation soft, \texttt{prompt\_start} & Qwen3.5--4B & 82.8 & 66.7 & 95.8 \\
Validation soft, \texttt{skill\_section} & Qwen3.5--4B & 84.9 & 77.8 & 95.8 \\
Validation soft, \texttt{prompt\_start} & Qwen3.5--9B & 83.3 & 77.8 & 100.0 \\
Validation soft, \texttt{skill\_section} & Qwen3.5--9B & 83.4 & 77.8 & \tbd{} \\
Validation soft, \texttt{prompt\_start} & Qwen3.6--35B-A3B & 87.3 & 94.4 & 99.0 \\
Validation soft, \texttt{skill\_section} & Qwen3.6--35B-A3B & 88.3 & 94.4 & \tbd{} \\
\bottomrule
\end{tabular}
}
\end{center}
\end{table}

\begin{table}[h]
\caption{Validation-selected checkpoint epochs in the original run collection. The new DocVQA skill-section runs are listed separately in Tables~\ref{tab:docvqa-skill-section-runs} and~\ref{tab:docvqa-skill-section-sweeps}. Each run evaluates checkpoints after epochs 1--3 and exports the checkpoint with the best validation score. Counts report how often each epoch is selected.}
\label{tab:validation-checkpoint-summary}
\begin{center}
\begin{tabular}{lrrrr}
\toprule
Task & Runs & Epoch 1 & Epoch 2 & Epoch 3 \\
\midrule
SearchQA & 13 & 3 & 2 & 8 \\
LiveMath & 13 & 2 & 5 & 6 \\
DocVQA & 13 & 7 & 5 & 1 \\
OfficeQA & 5 & 3 & 2 & 0 \\
Spreadsheet & 3 & 1 & 1 & 1 \\
ALFWorld & 6 & 3 & 1 & 2 \\
\midrule
Single-round QA total & 39 & 12 & 12 & 15 \\
Agentic total & 14 & 7 & 4 & 3 \\
\bottomrule
\end{tabular}
\end{center}
\end{table}



\begin{table}[h]
\caption{Skill-interpretability diagnostics. Skill cosine measures geometric similarity to the initialization, while greedy-decoded content match is a qualitative probe rather than evidence of semantic equivalence.}
\label{tab:interpretability-placeholder}
\begin{center}
\begin{tabular}{llcc}
\toprule
Task & Initialization & Skill cosine similarity & Greedy-decoded content match \\
\midrule
SearchQA & NL skill & 0.760 & Yes \\
LiveMath & NL skill & 0.934 & Yes \\
DocVQA & NL skill & 0.892 & Yes \\
OfficeQA & SkillOpt artifact & 0.907 & Yes \\
Spreadsheet & SkillOpt artifact & 0.754 & No \\
ALFWorld & SkillOpt artifact & 0.820 & Yes \\
\bottomrule
\end{tabular}
\end{center}
\end{table}

\begin{table}[h]
\caption{Full agentic grid under the default setting: Qwen3.6--35B-A3B, reproduced SkillOpt artifacts, and GPT-5.5-generated successful trajectories used as NTP supervision. Free-form generation; not comparable to Table~\ref{tab:alfworld-4b-forced}. For truncated skill rows, the corresponding untrained truncated skill baseline was not evaluated, so init-only scores and deltas are left blank rather than computed against a different source.}
\label{tab:agentic-ablation-grid}
\begin{center}
\resizebox{\linewidth}{!}{
\begin{tabular}{llllrrrrrr}
\toprule
Task & Source artifact & Skill construction & Skill length & Init-only val & Init-only test & Trained val & Trained test & $\Delta$ val & $\Delta$ test \\
\midrule
OfficeQA & initial.md & full artifact & auto & 25.0 & 26.2 & 16.7 & 11.4 & -8.3 & -14.8 \\
OfficeQA & final.md & first $n$ tokens & 8 & \tbd{} & \tbd{} & 33.3 & 36.0 & \tbd{} & \tbd{} \\
OfficeQA & final.md & first $n$ tokens & 32 & \tbd{} & \tbd{} & 41.7 & 33.8 & \tbd{} & \tbd{} \\
OfficeQA & final.md & first $n$ tokens & 256 & \tbd{} & \tbd{} & 37.5 & 33.7 & \tbd{} & \tbd{} \\
OfficeQA & final.md & full artifact & auto & 33.3 & 26.7 & 25.0 & 25.6 & -8.3 & -1.1 \\
\addlinespace
Spreadsheet & initial.md & full artifact & auto & 56.3 & 33.6 & 25.0 & 23.2 & -31.3 & -10.4 \\
Spreadsheet & final.md & first $n$ tokens & 8 & \tbd{} & \tbd{} & 50.0 & 28.2 & \tbd{} & \tbd{} \\
Spreadsheet & final.md & first $n$ tokens & 32 & \tbd{} & \tbd{} & 37.5 & 30.7 & \tbd{} & \tbd{} \\
Spreadsheet & final.md & first $n$ tokens & 256 & \tbd{} & \tbd{} & 37.5 & 30.0 & \tbd{} & \tbd{} \\
Spreadsheet & final.md & full artifact & auto & 18.8 & 52.5 & 37.5 & 35.7 & +18.7 & -16.8 \\
\addlinespace
ALFWorld & initial.md & full artifact & auto & 77.8 & 64.9 & 72.2 & 66.4 & -5.6 & +1.5 \\
ALFWorld & final.md & first $n$ tokens & 8 & \tbd{} & \tbd{} & 66.7 & 62.7 & \tbd{} & \tbd{} \\
ALFWorld & final.md & first $n$ tokens & 32 & \tbd{} & \tbd{} & 72.2 & 57.5 & \tbd{} & \tbd{} \\
ALFWorld & final.md & first $n$ tokens & 256 & \tbd{} & \tbd{} & 66.7 & 61.9 & \tbd{} & \tbd{} \\
ALFWorld & final.md & full artifact & auto & 83.3 & 79.1 & 83.3 & 71.6 & 0.0 & -7.5 \\
\bottomrule
\end{tabular}
}
\end{center}
\end{table}

\subsection{ALFWorld 4B Forced-Action Protocol}
\label{sec:appendix-alfworld-4b}

\begin{table}[!ht]
\caption{ALFWorld 4B forced-\actionopen{} success (\%) under semantics-preserving interface perturbations. The same trained skill block is evaluated without retraining on the locked 134-game test split. \xj{this table is a bit small to let it alone occupy an entire row; maybe consider wrapping it around text, or put two small tables/figures side by side in one row}}
\label{tab:alfworld-robustness}
\begin{center}
\small
\setlength{\tabcolsep}{6pt}
\begin{tabular}{lccc}
\toprule
Interface & Init & Trained & Gain \\
\midrule
Clean & 32.8 & \best{67.9} & \best{+35.1} \\
Paraphrase & 37.3 & \best{64.2} & +26.9 \\
Format & 22.4 & \best{62.7} & +40.3 \\
Irrelevant context & 29.9 & \best{61.9} & +32.1 \\
\bottomrule
\end{tabular}
\end{center}
\vspace{-0.8em}
\end{table}

\begin{table}[ht]
\caption{ALFWorld Qwen3.5--4B under the forced-\actionopen{} protocol. Same 294 teacher decisions and prompts; only the supervised token set and train-time rationale context change. Deltas are versus the untrained skill initialized from GPT-5.5 text.}
\label{tab:alfworld-4b-forced}
\begin{center}
\small
\resizebox{\linewidth}{!}{
\begin{tabular}{llccc}
\toprule
Condition & Train-time rationale & Best epoch & Val (n=18) & Test (n=134) \\
\midrule
Untrained skill block & --- & 0 & 4/18 & \acc{32.8}{+0.0} \\
Full-response supervision & generated as target & 3 & 14/18 & \best{\acc{67.9}{+35.1}} \\
Action-only supervision & absent & 2 & 14/18 & \best{\acc{67.9}{+35.1}} \\
Masked-rationale supervision & gold think, no loss & 1 & 7/18 & \acc{46.3}{+13.5} \\
\bottomrule
\end{tabular}
}
\end{center}
\vspace{-0.6em}
\end{table}

Table~\ref{tab:alfworld-4b-forced} uses Qwen3.5--4B, GPT-5.5 skill initialization at \texttt{skill\_section}, auto skill block length 2806, 31 successful teacher trajectories (294 per-step examples), learning rate $10^{-4}$, three epochs, and a validation gate on 18 games. The decoder appends a literal \actionopen{} to the chat prompt and stops at \actionclose{} or EOS with a 512-token cap. This protocol is not comparable to 4B free-form initialization ($\approx 89/134$) or to the 35B free-form table.

Full-response and masked-rationale runs train on the stored teacher reply (rationale span plus \actionopen{}\ldots\actionclose{}), with mean 66.4 target tokens. Action-only training strips rationale from the sequence and completes a forced \actionopen{} prefix (mean 9.5 tokens), matching evaluation. Masked-rationale keeps the full sequence but applies loss only from \actionopen{} onward (mean 13.5 supervised tokens, 52.9 masked). Table~\ref{tab:alfworld-4b-paired} reports pairwise McNemar tests on the 134 test games. Full-response versus action-only is a 13/13 split (exact $p=1$): the two skill blocks match in aggregate success and are not the same policy.

\begin{table}[h]
\caption{Paired tests on ALFWorld 4B forced-action rollouts ($n=134$). Full-response and action-only match in count; masked-rationale is worse on both sides.}
\label{tab:alfworld-4b-paired}
\begin{center}
\resizebox{\linewidth}{!}{
\begin{tabular}{lccc}
\toprule
Contrast & $\Delta$ (pp) & McNemar $n_{01}/n_{10}$ & Exact $p$ \\
\midrule
Full-response vs action-only & $0.0$ & 13 / 13 & $1.0$ \\
Full-response vs masked-rationale & $-21.6$ & 7 / 36 & $9\times 10^{-6}$ \\
Action-only vs masked-rationale & $-21.6$ & 3 / 32 & $4\times 10^{-7}$ \\
\bottomrule
\end{tabular}
}
\end{center}
\end{table}

Masked-rationale is a train--test mismatch diagnostic, not an action-only baseline. Training predicts $p(a\mid h,r_{\mathrm{gold}})$ while forced-action evaluation predicts $p(a\mid h)$. Action NLL falls from $5.7\times 10^{-3}$ at epoch 1 to $6.6\times 10^{-5}$ at epoch 3; validation never exceeds $7/18$. Masking rationale tokens from the loss does not remove their causal role as teacher-forced inputs.

\paragraph{Interface perturbations.}
We evaluate the action-only 4B checkpoint from Table~\ref{tab:alfworld-4b-forced} without retraining, under frozen templates that paraphrase instructions, rewrap formatting, or insert a fixed task-irrelevant context block. Environments, legal actions, the 134 test episodes, \texttt{skill\_section} insertion, and forced-\actionopen{} decoding are held fixed. Table~\ref{tab:alfworld-robustness} reports percentages. Table~\ref{tab:alfworld-robustness-paired} reports paired transitions. Formatting changes drop initialization from $32.8\%$ to $22.4\%$ while the trained skill block remains at $62.7\%$; we do not interpret this single perturbation as improved format robustness, only as the absence of stronger format sensitivity than initialization.

\begin{table}[h]
\caption{Paired ALFWorld 4B tests of initialization versus the trained skill block under each interface ($n=134$). $n_{01}$ is init fail $\to$ trained success; $n_{10}$ is the reverse. Bootstrap CIs use 10{,}000 resamples (seed 1).}
\label{tab:alfworld-robustness-paired}
\begin{center}
\small
\begin{tabular}{lcccc}
\toprule
Interface & $n_{01}/n_{10}$ & $\Delta$ (pp) & Exact $p$ & 95\% CI \\
\midrule
Clean & $48/1$ & $+35.1$ & $1.8\times 10^{-13}$ & $[26.9,\,43.3]$ \\
Paraphrase & $38/2$ & $+26.9$ & $1.5\times 10^{-9}$ & $[19.4,\,35.1]$ \\
Format & $58/4$ & $+40.3$ & $2.6\times 10^{-13}$ & $[31.3,\,50.0]$ \\
Irrelevant context & $45/2$ & $+32.1$ & $1.6\times 10^{-11}$ & $[23.9,\,40.3]$ \\
\bottomrule
\end{tabular}
\end{center}
\end{table}

\FloatBarrier
\phantomsection
\paragraph{Trajectory behavior.}
\label{sec:appendix-alfworld-behavior}
The $44/134{\to}91/134$ success change does not by itself say what the agent does differently. We therefore score the saved forced-\actionopen{} test conversations for the same action-only skill block used in the clean robustness condition, paired against initialization on the locked 134 episodes. No new training or rollout was required: initialization test conversations come from the robustness clean dump, and trained traces are identical to that dump's trained conversations on every episode.

An action is \emph{environment-rejected} when the environment reply is exactly \nothinghappens{}. This is not a missing-tag or parser failure; forced-\actionopen{} decoding already requires an action completion. A \emph{loop} episode contains an $A{\to}B{\to}A$ action cycle or a run of three or more identical actions. \emph{Recovery} is scored on episodes with at least one rejected or malformed action: after such a step, a later step emits a different action whose reply is not \nothinghappens{}. Immediate-repeat rate is the episode mean of consecutive identical actions. LLM labeling was not used.

Table~\ref{tab:alfworld-behavior-full} reports the full paired summary. Success is one-sided: 48 initialization failures become successes, 43 episodes succeed in both conditions, 42 fail in both, and one previously successful episode regresses ($n_{01}/n_{10}=48/1$; Table~\ref{tab:alfworld-robustness-paired}). Median length falls from 50 to 8 because initialization often stalls: $90/134$ initialization episodes hit the 50-step horizon versus $43/134$ after training, and those 43 trained cap hits are exactly the remaining failures. Mean rejected actions fall from 5.17 to 2.46, immediate repeats from $0.093$ to $0.015$, and loop episodes from $0.478$ to $0.216$. Recovery is already high ($0.955$ on 67 initialization episodes with a rejection, versus $0.971$ on 35 trained episodes), so it does not explain the gain.

The most informative subset is the 48 fail$\to$success episodes. These are the same tasks under both skill blocks: initialization hits the horizon in every case, while the trained skill block succeeds. Trained traces use 39.4 fewer steps on average, incur 7.10 fewer rejected actions, and drop loop incidence from $0.646$ to $0.042$. Shorter successful traces are therefore not only a mechanical consequence of having more successes in the overall mean.

Together with the loss-support ablation, this suggests that supervising successful actions is reflected in downstream action dynamics, not merely in output-format compliance. These statistics associate the trained skill block with measurable sequential-action changes. They do not show that \method{} learned a general agent policy, and they are not a causal isolation of any one failure mode: success, length, rejections, and loops move together.

\begin{table}[h]
\caption{Paired ALFWorld 4B action-behavior summary ($n{=}134$). Paired change is trained minus initialization. Recovery denominators are episodes with at least one rejected or malformed action.}
\label{tab:alfworld-behavior-full}
\begin{center}
\small
\begin{tabular}{lccc}
\toprule
Metric & Init & Trained & Paired change \\
\midrule
Success & $44/134$ ($32.8\%$) & $91/134$ ($67.9\%$) & $+47/134$ \\
Median steps & 50 & 8 & $-42$ \\
Rejected actions / episode & 5.17 & 2.46 & $-2.71$ \\
Immediate-repeat rate & 0.093 & 0.015 & $-0.077$ \\
Loop rate & 0.478 & 0.216 & $-0.261$ \\
Recovery rate & 0.955 & 0.971 & $+0.016$ \\
\bottomrule
\end{tabular}
\end{center}
\end{table}

We select four exemplars using fixed criteria: largest length reduction among fail$\to$success (\texttt{test\_0013}); largest drop in immediate repeats among fail$\to$success traces that had repeats (\texttt{test\_0075}); largest drop in \nothinghappens{} among fail$\to$success (\texttt{test\_0018}); and the unique success$\to$fail regression (\texttt{test\_0118}). Representative trajectory excerpts: initialization repeatedly issues \texttt{examine mug} instead of taking it, repeats \texttt{cool bread} after picking up the wrong object, or attempts \texttt{take mug from desk 1} while the mug is on shelf 2; the trained skill block avoids those loops. The regression is a pick-two CDs task in which initialization finishes in 14 steps while the trained skill block wanders among shelves until the horizon.

\begin{table}[h]
\caption{Side-by-side trajectory excerpts from the four deterministic exemplars. Remaining steps continue the same pattern until success or the 50-step cap.}
\label{tab:alfworld-behavior-exemplars}
\begin{center}
\scriptsize
\setlength{\tabcolsep}{2.5pt}
\resizebox{\linewidth}{!}{
\begin{tabular}{cllll}
\toprule
Step & Init action & Init reply & Trained action & Trained reply \\
\midrule
\multicolumn{5}{l}{\texttt{test\_0013}: examine mug with desklamp; init 50 steps fail $\to$ trained 3 steps success} \\
0 & go to desk 1 & arrive; mug and desklamp visible & go to desk 1 & arrive; mug and desklamp visible \\
1 & use desklamp 1 & lamp on & take mug 1 from desk 1 & picked up \\
2 & examine mug 1 & \nothinghappens{} & use desklamp 1 & success \\
3--49 & examine / use lamp loop & \nothinghappens{} / no progress & --- & --- \\
\addlinespace
\multicolumn{5}{l}{\texttt{test\_0075}: cool bread, put on countertop; init 50 fail $\to$ trained 14 success} \\
0--8 & search; take apple from microwave & mixed & search countertops & bread found at step 9 \\
9 & cool apple 1 with fridge 1 & cooled apple & take bread 1 from countertop 3 & picked up \\
10--49 & cool bread 1 with fridge 1 & \nothinghappens{} (repeat) & cool bread; place on countertop 3 & success at step 13 \\
\addlinespace
\multicolumn{5}{l}{\texttt{test\_0018}: put a mug in desk; init 50 fail $\to$ trained 7 success} \\
0--3 & visit desks/shelf 1 & no mug & visit desks; go to shelf 2 & mug 1 visible \\
4--11 & take mug 1 from desk 1 & \nothinghappens{} (repeat) & take mug 1 from shelf 2 & picked up \\
12--49 & revisit desks; retry take from desk 1 & \nothinghappens{} & move mug 1 to desk 1 & success at step 6 \\
\addlinespace
\multicolumn{5}{l}{\texttt{test\_0118}: find two CDs, put in safe; init 14 success $\to$ trained 50 fail} \\
0--6 & open drawers; take cd 3 from drawer 4 & CD in hand & visit shelf 1--7 & no CDs; some \nothinghappens{} \\
7--13 & place in safe; take cd 1 from desk 1; place & success & repeat shelf tour & horizon, unsolved \\
\bottomrule
\end{tabular}
}
\end{center}
\end{table}

\FloatBarrier

\paragraph{Prompt-mismatched reconstruction.}
An earlier action-BC run that rebuilt user prompts with a full interaction history and a different step-0 template scored $71/134$ (val $13/18$). Example ids and extracted actions match the 294 action-only rows; 201/294 training prompts use a history window longer than the two most recent entries exposed at evaluation. We treat that run as a development artifact, not as evidence against action-only supervision, and we do not train a matched-history reconstruction.

\paragraph{Preference, orthogonal axis.}
Pairwise preference asks a different question: whether relative outcome feedback can train the skill block when a state has no unique canonical action. Candidates are sampled and continued with a frozen policy. Verified success gaps yield $a_+\succ a_-$, scored by mean token log-probability $s(\cdot)$:
\begin{equation}
\mathcal{L}_{\mathrm{pref}}(\Delta p)
=
-\log\sigma\bigl(\beta(s(a_+)-s(a_-))\bigr)
+\lambda\bigl(-s(a_+)\bigr).
\end{equation}
One seed on 81 training pairs ($\beta=1$, $\lambda=0.1$) reaches $82/134$ (val $12/18$ at epoch 1) versus initialization $44/134$ (McNemar $43/5$, $p=1.4\times 10^{-8}$). A second optimization seed, holding the pair-generation seed fixed, scored $3/18$ on validation and kept the untrained skill block. We do not rank $82/134$ against $91/134$, do not run a third seed, and do not extend preference to OfficeQA, Spreadsheet, SearchQA, LiveMath, or DocVQA.

\subsection{ALFWorld learned-update geometry and layer readout}
\label{sec:appendix-alfworld-mechanism}
\xj{New appendix for the 2026-09-26 experiments. Geometry jobs 139434/139435 (16-game pilot) and 139483 (134-game confirmation). Layer jobs 139427/139428 (coarse scores), 139433 (134-game rollout), and 139484 (64-decision resplit). No new SoftSkill training. Please verify every count against \texttt{outputs/agentic/analysis/softskill\_prefix\_geometry/} and \texttt{outputs/agentic/analysis/softskill\_layer\_causality/alfworld\_rollout\_139433/} before submission.}

Both analyses use the same validation-selected action-only skill block as Table~\ref{tab:alfworld-4b-forced}: Qwen3.5--4B, GPT-5.5 skill initialization at \texttt{skill\_section}, length $2{,}806$, and the locked 134-game \texttt{valid\_unseen} split. Geometry and the published $44/134{\to}91/134$ numbers use the vLLM prompt-embeds path. The layer rollout uses local HuggingFace greedy decoding of the same checkpoint, where initialization and the trained skill score $48/134$ and $94/134$. Layer comparisons use those HuggingFace endpoints. Low-rank reconstructions and per-position norm-matched random directions were not evaluated.\xj{Confirm the checkpoint identity: geometry log names \texttt{outputs/agentic/trajectory\_sft\_baseline/full\_137105/best\_prefix.pt}, epoch 2, val $14/18$, SHA256 \texttt{ee6638569c0fe2f084210a0595e94a4cb506e22753a8ba55af322f3715caa22c}. \texttt{latest\_prefix.pt} is the rejected epoch 3 and was not used. Later notes identify job 137105 as action-only, not full-response (job 137358). Reconstructed $\lVert p_{\mathrm{init}}\rVert_F=36.637$ versus the training log $36.639$; $\lVert\Delta\rVert_F=4.000$ versus $4.001$.}

\paragraph{Displacement geometry.}
Let $\Delta = p_{\mathrm{trained}} - p_{\mathrm{init}}$. Table~\ref{tab:alfworld-delta-geometry} summarizes the learned offset. Every position moves, but the largest $10\%$ of positions account for only $19\%$ of the squared displacement, so the update is not a few critical soft tokens. Nearest-token identity is unchanged under both cosine and $\ell_2$ at all $2{,}806$ positions, which places $\Delta$ inside the Voronoi cell of the original skill tokens. That identity is not a semantic decoding of the skill. The singular spectrum is wide, so $\Delta$ is also not a low-rank steering direction: $1{,}141$ and $1{,}937$ components are required for $90\%$ and $99\%$ of the Frobenius energy.

\begin{table}[h]
\caption{Geometry of the learned ALFWorld displacement on the action-only Qwen3.5--4B skill. Nearest-token probes use the frozen embedding table and are not a semantic readout.}
\label{tab:alfworld-delta-geometry}
\begin{center}
\small
\begin{tabular}{lc}
\toprule
Quantity & Value \\
\midrule
$\lVert\Delta\rVert_F / \lVert p_{\mathrm{init}}\rVert_F$ & $0.109$ \\
Per-position $\lVert\Delta_i\rVert_2$ mean / median / max / min & $0.074$ / $0.072$ / $0.217$ / $0.034$ \\
Mean cosine with initialization & $0.994$ \\
Nearest-token identity changes (cosine and $\ell_2$) & $0/2{,}806$ \\
Leading singular-component energy & $3.9\%$ \\
Components for $90\%$ / $99\%$ energy & $1{,}141$ / $1{,}937$ \\
Squared-displacement share of the top $10\%$ positions & $19\%$ \\
\bottomrule
\end{tabular}
\end{center}
\end{table}

\paragraph{Continuous path, random control, and vocabulary snap.}
The interpolated skill is $p(\alpha)=p_{\mathrm{init}}+\alpha\Delta$. The random control is an isotropic Gaussian tensor of the same shape, rescaled to $\lVert\Delta\rVert_F$ and added to the initialization. The snap replaces each position by its cosine-nearest vocabulary embedding. A fixed 16-episode slice, \texttt{test\_0000}--\texttt{test\_0015}, has initialization $3/16$ and trained $12/16$. On that slice, $\alpha\in\{0.25,0.50,0.75\}$ scores $4/16$, $12/16$, and $13/16$; random seeds $0$, $1$, and $2$ score $4/16$, $2/16$, and $2/16$; snapping either endpoint scores $3/16$. The $\alpha=0.50$ successes cover $11$ of the $12$ trained successes. The curve jumps between $0.25$ and $0.50$ and is not pointwise monotone. The 134-game confirmation therefore covers only the three conditions that decide the claim: $\alpha=0.50$, the strongest pilot random seed, and the trained snap (Table~\ref{tab:alfworld-geometry-paired}).\xj{Pilot-only conditions: $\alpha=0.25$, $\alpha=0.75$, random seeds 1 and 2, and snap(init). Do not promote them to the same status as the 134-game rows. The full-test random control is a single direction, seed 0, not an average over directions.}

\begin{table}[h]
\caption{Paired ALFWorld tests of continuous and discrete controls on the vLLM path ($n=134$). Deltas are the named condition minus initialization ($44/134$). $n_{01}$ is initialization failure and condition success; $n_{10}$ is the reverse. Bootstrap CIs use $10{,}000$ resamples (seed 1).}
\label{tab:alfworld-geometry-paired}
\begin{center}
\small
\resizebox{\linewidth}{!}{
\begin{tabular}{lccccc}
\toprule
Condition & Success & $\Delta$ & $n_{01}/n_{10}$ & Exact $p$ & 95\% CI \\
\midrule
Trained & $91/134$ & $+0.351$ & $48/1$ & $1.8\times 10^{-13}$ & $[0.269,\,0.433]$ \\
$\alpha=0.50$ & $87/134$ & $+0.321$ & $43/0$ & $2.3\times 10^{-13}$ & $[0.246,\,0.403]$ \\
Random, seed 0 & $49/134$ & $+0.037$ & $8/3$ & $0.23$ & $[-0.007,\,0.090]$ \\
Snap(trained) & $44/134$ & $0$ & $0/0$ & $1$ & $[0,\,0]$ \\
\bottomrule
\end{tabular}
}
\end{center}
\end{table}

Against the trained skill, $\alpha=0.50$ differs by $-0.030$ (95\% CI $[-0.090,\,0.030]$; trained-only / condition-only successes $10/6$; $p=0.45$). The random control differs by $-0.313$ ($[-0.396,\,-0.231]$; $44/2$; $p=3.1\times 10^{-11}$). Snap(trained) matches initialization on every episode, so its contrast with the trained skill repeats the initialization row ($48/1$; $p=1.8\times 10^{-13}$). Halfway interpolation does not lose any episode that initialization solves. These tests do not establish that $\alpha=0.50$ and the trained skill are the same policy.\xj{Orientation check: this $n_{01}/n_{10}$ matches Table~\ref{tab:alfworld-robustness-paired}. The trained-versus-$\alpha$ counts $10/6$ are trained-only / $\alpha$-only, the opposite order. $\alpha=0.50$ loses no initialization success ($n_{10}=0$ in the table above).}

\paragraph{Layer patch.}
For a selected set of decoder layers, two forwards differ only in the skill-block embedding. At those layers, the trained forward's hidden state at skill-block positions is replaced by the initialization forward's hidden state. Later layers run as usual. Task-token states are not written. Qwen3.5--4B has 32 decoder layers. Decode steps shorter than the skill-block span are left unpatched, because the skill-block state is already in the cache from prefill. Table~\ref{tab:alfworld-layer-patch} reports environment success. Patching layers $0$--$7$ drops success from $94/134$ to $38/134$, ten episodes below the $48/134$ initialization on this rollout. That deficit is not a significant difference ($p=0.052$; $\Delta=-0.075$; 95\% CI $[-0.142,\,-0.007]$). Patching layers $8$--$15$ removes about three fifths of the gain. Patching layers $24$--$31$ is unresolved from the trained skill.

\begin{table}[h]
\caption{HuggingFace greedy ALFWorld rollout under prefix-state patches ($n=134$). Deltas, discordant counts, and CIs are versus the trained skill on this same rollout ($94/134$), not versus the vLLM $91/134$. ``Trained only / condition only'' counts episodes solved by exactly one side. Bootstrap CIs use $10{,}000$ resamples (seed 1).}
\label{tab:alfworld-layer-patch}
\begin{center}
\small
\resizebox{\linewidth}{!}{
\begin{tabular}{lccccc}
\toprule
Condition & Success & $\Delta$ vs.\ trained & Trained only / condition only & Exact $p$ & 95\% CI \\
\midrule
Initialization & $48/134$ & $-0.343$ & $49/3$ & $1.0\times 10^{-11}$ & $[-0.433,\,-0.254]$ \\
Patch layers $0$--$7$ & $38/134$ & $-0.418$ & $57/1$ & $4.1\times 10^{-16}$ & $[-0.500,\,-0.336]$ \\
Patch layers $8$--$15$ & $66/134$ & $-0.209$ & $31/3$ & $7.7\times 10^{-7}$ & $[-0.284,\,-0.134]$ \\
Patch layers $24$--$31$ & $90/134$ & $-0.030$ & $5/1$ & $0.22$ & $[-0.067,\,0.000]$ \\
\bottomrule
\end{tabular}
}
\end{center}
\end{table}

A teacher-forced probe scores mean token log probability of reference actions, which is not the 134-game success vector. On 16 training decisions, initialization scores $-0.182$ and the trained skill $-0.039$; patching layers $0$--$7$ returns $-0.192$, while layers $16$--$23$ and $24$--$31$ stay at $-0.042$ and $-0.038$. On 64 training decisions, initialization scores $-0.192$ and the trained skill $-0.050$. Patching layers $0$--$3$, $4$--$7$, $8$--$11$, and $12$--$15$ removes $100\%$, $77\%$, $44\%$, and $15\%$ of that gap. The early representation gap is not the largest one: the skill-block hidden-state difference grows from about $438$ in the first coarse band to about $1{,}067$ in the last, while the behavioral effect shrinks. The patch is a single state replacement, and later layers could in principle rewrite it; layers $0$--$3$ already account for the full log-probability gap. Bands $16$--$31$ were not swept layer by layer after the environment signal there disappeared.\xj{Log-probability rows are training decisions, not the 134 test games. The $100/77/44/15$ percentages are the share of the trained-minus-init gap removed, not accuracy. SearchQA: on a 32-question validation slice the trained skill's answer log probability was worse than initialization ($-4.69$ vs.\ $-4.07$), so no SearchQA layer localization was run. Do not write the early-layer claim as cross-task.}

\subsection{Skill Concatenation Diagnostic}
\label{sec:appendix-composition}
\xj{Appendix composition details. Prefer cutting the main-text concatenation paragraph before cutting this section, or drop both together if a later revision recenters away from composition. Do not promote these numbers into a claim that learned subskills intrinsically interfere.}

This appendix records a concatenation diagnostic on the 4B forced-\actionopen{} ALFWorld protocol, not a paper composition result. Two operational rules were taken from the ALFWorld skill: $A$ is transform-before-delivery (clean/heat/cool before placing) and $B$ is direct delivery after the object is in hand. Independently optimized length-32 skill blocks were trained only on $A$-only or $B$-only decisions and concatenated at the same \texttt{skill\_section} slot at inference time. The metric is greedy teacher-action match, not full-game success.

The construction is not a strict composition test. $A$ fit its 34 training decisions ($25{\to}30/34$) but did not improve a 13-row $A$-only validation set ($4/13{\to}3/13$), so validation selected initialization. $B$ saturated its 55 $B$-only training decisions ($50{\to}55/55$) and had no uncontaminated $B$-only holdout. Example ids are disjoint across $A$-only, $B$-only, and $A{+}B$, but episodes are not: all 18 $A{+}B$ episodes also appear in both component-training games. The missing factorial geometry is therefore not merely ``too little ALFWorld data''; $A$-only / $B$-only / $A{+}B$ support cannot be cut cleanly from existing episodes.

Table~\ref{tab:composition-concat} reports the $A{+}B$ bucket ($n=31$). $B$ alone is strongest ($23/31$). All-zero skill blocks of length 32 largely preserve $B$ ($23/31$ and $22/31$). Duplicate and random skill blocks of the same length drop to $21/31$ and $19$--$20/31$. Concatenations containing $A$, including untrained $A_{\mathrm{init}}$, score $17$--$19/31$. Zero padding of the extra 32 positions is therefore not equivalent to inserting another independently optimized skill block. We interpret this as evidence against a mere length/OOD-position failure mode, not as proof that two validated subskills interfere.

\begin{table}[h]
\caption{ALFWorld 4B forced-\actionopen{} concatenation diagnostic on $A{+}B$ decisions ($n=31$). Scores are greedy teacher-action matches. $B$ is the epoch-3 $B$-only skill block; $\mathrm{ZERO}$ is an all-zero length-32 tensor; $\mathrm{RANDOM}$ is Gaussian noise rescaled to $B$'s mean per-token $\ell_2$. This is not a paper composition claim.}
\label{tab:composition-concat}
\begin{center}
\small
\begin{tabular}{lc}
\toprule
Skill & $A{+}B$ match \\
\midrule
$B$ & $23/31$ \\
$[\mathrm{ZERO};B]$ & $23/31$ \\
$[B;\mathrm{ZERO}]$ & $22/31$ \\
$[B;B]$ & $21/31$ \\
$[B;\mathrm{RANDOM}]$ & $20/31$ \\
$[\mathrm{RANDOM};B]$ & $19/31$ \\
$A$-containing concatenations & $17$--$19/31$ \\
$A_{\mathrm{init}}$ / $A_{\mathrm{ep3}}$ alone & $15/31$ / $14/31$ \\
\bottomrule
\end{tabular}
\end{center}
\end{table}

SearchQA was audited as a fallback and is a \emph{no-go} for the same experiment. The skill's semantically meaningful operations---kinship inversion, work-versus-role, parent-class induction---almost never co-occur as $A$-only / $B$-only / $A{+}B$ in the paper split, and expanding to the full SearchQA dump does not create a clean factorial geometry (kinship $\cap$ work-versus-role is 60 examples before any holdout cut). Pairs with enough counts are surface templates, such as type deixis (\texttt{this \textless type\textgreater}) versus year/superlative cues. Training skill blocks on those labels would manufacture a lexical-split experiment rather than test reusable operations. About $43\%$ of the paper-split items are also Jeopardy-dump shortcuts. \method{}'s training machinery is compatible with SearchQA; the task structure is not.

\begin{figure}[t]
\begin{center}
\includegraphics[width=0.33\linewidth]{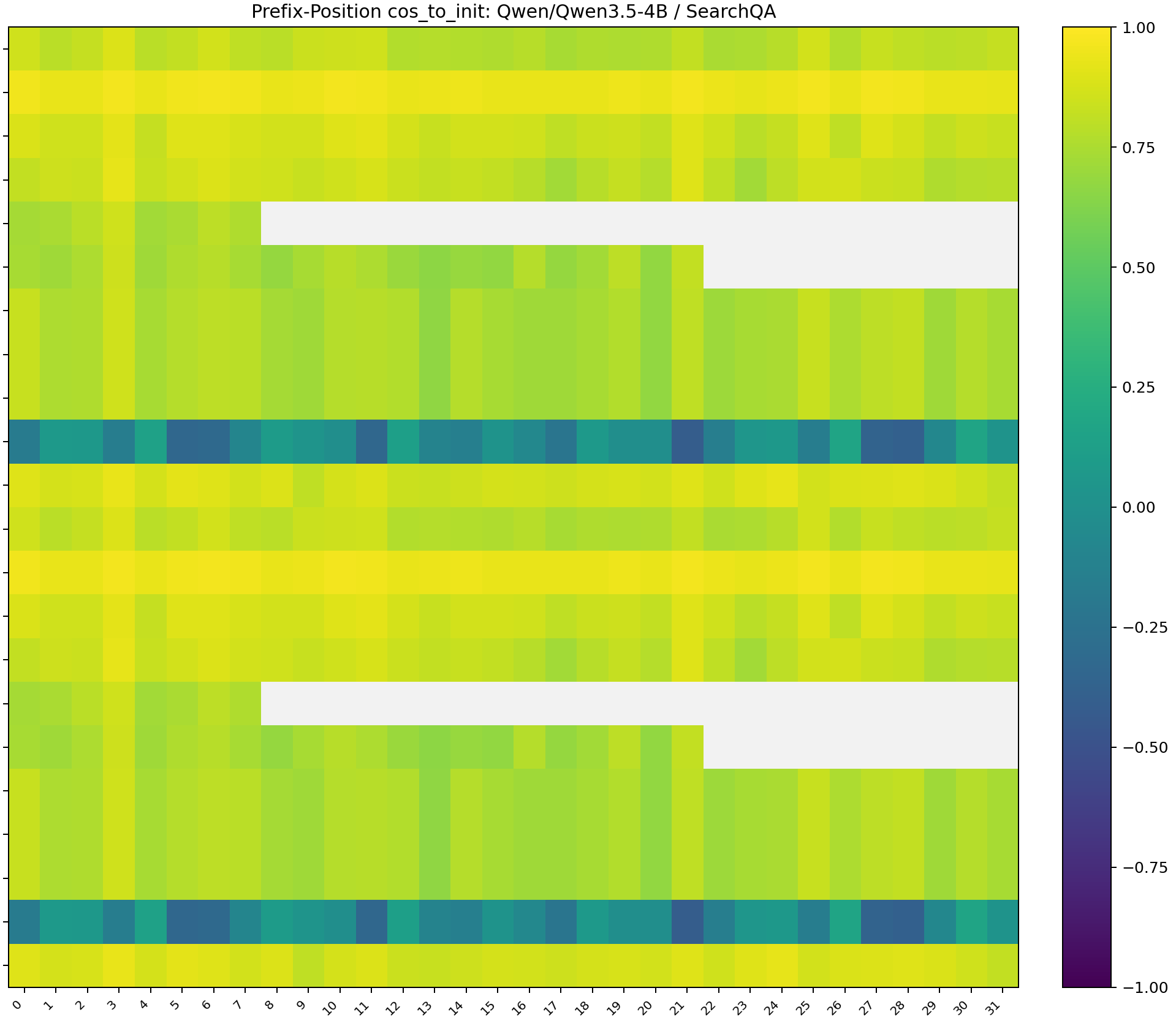}
\hfill
\includegraphics[width=0.30\linewidth]{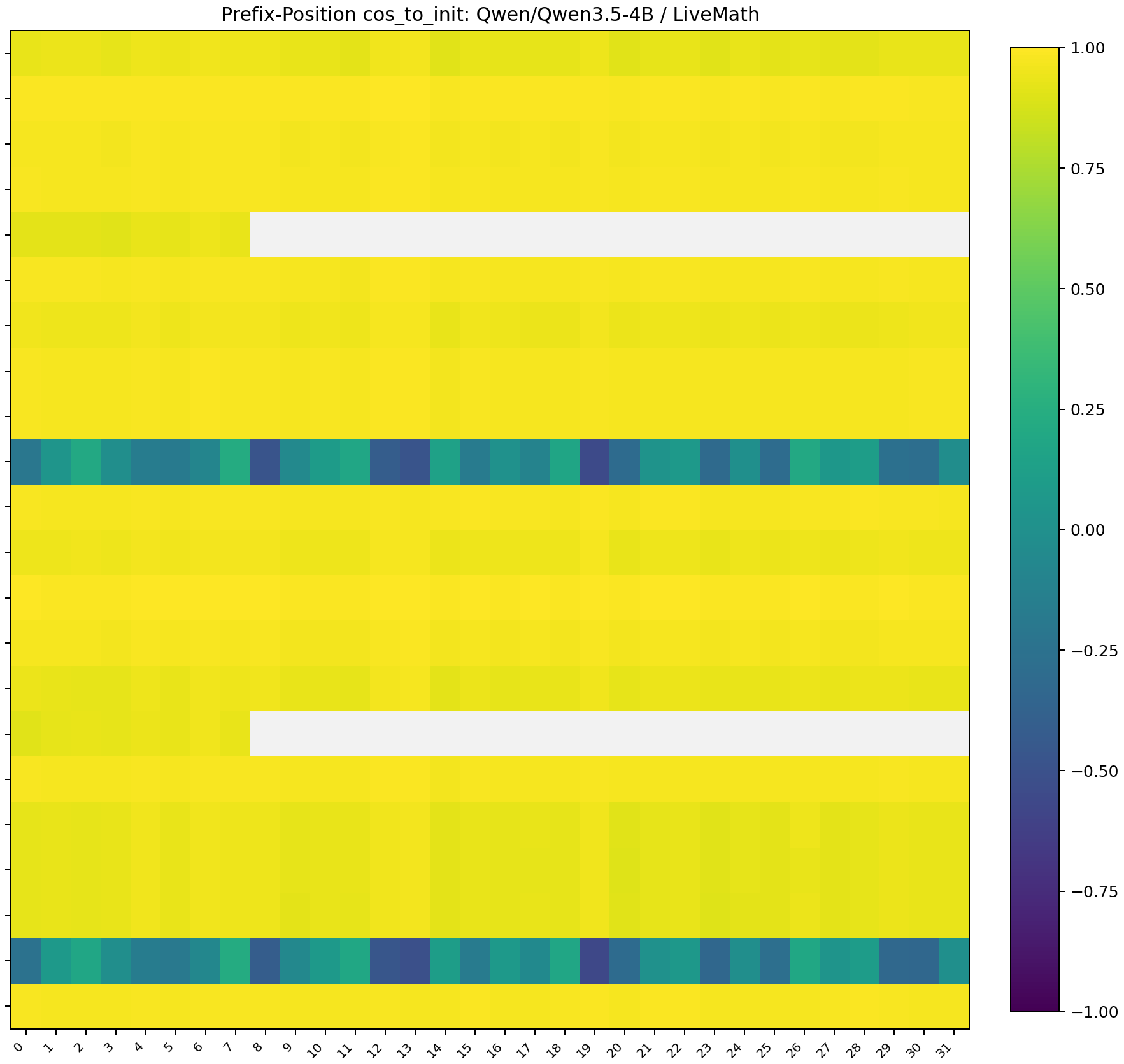}
\hfill
\includegraphics[width=0.33\linewidth]{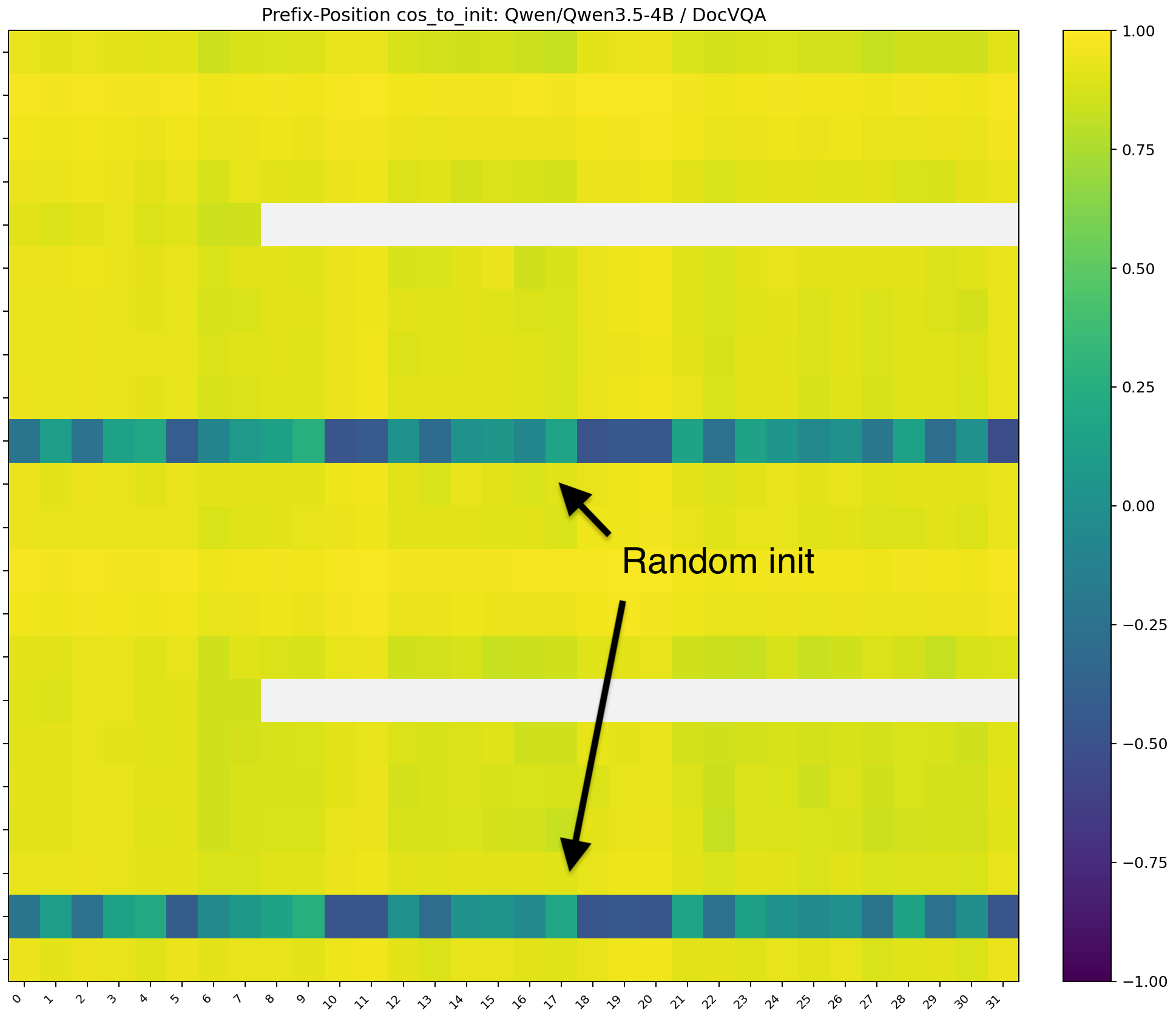}
\end{center}
\caption{Per-position cosine similarity between each learned Qwen3.5--4B soft skill and its initialization, ordered as SearchQA, LiveMath, and DocVQA. Higher values indicate a smaller learned displacement from the initial embedding sequence. For DocVQA, the earlier runs shown here and logged under \texttt{skill\_section} are effective prompt-start insertions in the vision-language path. SearchQA shows larger divergence from initialization than the other QA tasks, and mean-pooled initialization requires the largest learned displacement.}
\label{fig:prefix-cos-to-init-4b}
\end{figure}

\end{document}